\documentclass{article}
\usepackage{ijcai21}

\usepackage{times}
\usepackage{soul}
\usepackage{url}
\usepackage[utf8]{inputenc}
\usepackage[small]{caption}
\usepackage{graphicx}
\usepackage{amsmath}
\usepackage{amsthm}
\usepackage{booktabs}
\usepackage{algorithm}
\usepackage{algorithmic}
\usepackage{subfigure}
\usepackage{graphicx}
\usepackage{color}
\usepackage{mathbbol}
\usepackage{listings}
\usepackage{xcolor}

\definecolor{codegreen}{rgb}{0,0.6,0}
\definecolor{codegray}{rgb}{0.5,0.5,0.5}
\definecolor{codepurple}{rgb}{0.58,0,0.82}
\definecolor{backcolour}{rgb}{0.95,0.95,0.92}

\lstdefinestyle{mystyle}{
	backgroundcolor=\color{backcolour},   
	commentstyle=\color{codegreen},
	keywordstyle=\color{magenta},
	numberstyle=\tiny\color{codegray},
	stringstyle=\color{codepurple},
	basicstyle=\ttfamily\footnotesize,
	breakatwhitespace=false,         
	breaklines=true,                 
	captionpos=b,                    
	keepspaces=true,                            
	numbersep=5pt,                  
	showspaces=false,                
	showstringspaces=false,
	showtabs=false,                  
	tabsize=2
}

\newtheorem{theorem}{Theorem}

\title{AdaL: Adaptive Gradient Transformation Contributes to Convergences and Generalizations}

\iftrue
\author{
	Hongwei Zhang$^{1}$\footnote{These authors contributed equally to this work.} \and 
	Weidong Zou$^{1*}$\and
	Hongbo Zhao$^{1*}$\and
	Qi Ming$^1$\and
	Tijin Yan$^1$\and \\
	Yuanqing Xia$^1$\and
	Weipeng Cao$^2$\\
	\affiliations
	$^1$Beijing Institute of Technology\\
	$^2$Shenzhen University\\
}
\fi

\begin{document}

\maketitle

\begin{abstract}
Adaptive optimization methods have been widely used in deep learning. They scale the learning rates adaptively according to the past gradient, which has been shown to be effective to accelerate the convergence. However, they suffer from poor generalization performance compared with SGD. Recent studies point that smoothing exponential gradient noise leads to generalization degeneration phenomenon. Inspired by this, we propose AdaL, with a transformation on the original gradient. AdaL accelerates the convergence by amplifying the gradient in the early stage, as well as dampens the oscillation and stabilizes the optimization by shrinking the gradient later. Such modification alleviates the smoothness of gradient noise, which produces better generalization performance. We have theoretically proved the convergence of AdaL and demonstrated its effectiveness on several benchmarks.
\end{abstract}

\section{Introduction}

Deep learning has shown great potential in many fields, especially in computer vision and natural language processing. As an important role in deep learning, optimization algorithm has attracted more and more attention. In recent years, researchers
have proposed various optimizers (also known as optimization algorithms) for deep learning, which can be categorized into two branches: one is the stochastic gradient decent and its variants, such as SGD with moment \cite{sutskever2013importance}, Nesterov accelerated gradient (NAG) \cite{nesterov1983method}, and the other is adaptive methods, such as AdaGrad \cite{duchi2011adaptive}, RMSProp \cite{tieleman2012lecture}, Adam \cite{Kingma2014adam}, AMSGrad \cite{Reddi2018amsgrad}, NosAdam \cite{huang2019nostalgic}, etc.. These algorithms have the following generic  framework:
\begin{equation}
	\begin{aligned}
		x_{t+1}=x_t -\frac{\alpha_t}{\psi(g_1, ..., g_t)} \phi(g_1, ..., g_t)
	\end{aligned}
\end{equation}
where $g_t$ is the gradient of the $i$-th time step,
$\alpha_t/\psi(\cdot) $ is the adaptive learning rate, and $\phi(\cdot)$ is the gradient estimation. Different $\psi(\cdot)$ and $\phi(\cdot)$ derive different optimization algorithms. For example, in SGD, $\psi(\cdot)=I$ and $\phi(\cdot)=g_t$, while in Adam, $\phi(\cdot)=(1-\beta_1)\sum_{i=1}^{t}\beta_1 (t-i)g_i$ and $\psi(\cdot)=(1-\beta_2)\text{diag}(\sum_{i=1}^{t}\beta_2 (t-i)g_i^2)$. 

The key idea of SGD is to randomly select a mini-batch of samples  to compute the gradients and update the parameters, thereby reducing the computational cost of the batch gradient descent from $O(n)$ to $O(1)$. To accelerate the convergence speed of SGD, SGD with momentum \cite{sutskever2013importance} has been proposed to reduce the oscillation of SGD. Based on SGDM, Nesterov accelerated gradient \cite{nesterov1983method} uses $x_t-\gamma m_{t-1}$ to replace $x_t$ to give us a prediction of the next position of the parameter. 

Adaptive algorithms \cite{duchi2011adaptive,zeiler2012adadelta,Kingma2014adam,loshchilov2018decoupled,liu2019variance,li2020adax,zhuang2020adabelief} are proposed to solve the problem that SGD scales the gradient uniformly in all directions, which may lead to limited training speed as well as poor performance when the training data are sparse. Adam \cite{Kingma2014adam} is perhaps the most popular adaptive stochastic optimization method which uses exponential moving average to estimate the learning rate scheduler and gradient adaptively. Compared with SGD, Adam adjusts the learning rate according to the gradient value of the independent variable in each dimension. Such modification contributes to the convergence speed in the training phase. However, adaptive gradient algorithms usually suffer from poor generalization performance while SGD performs better. 


There are several studies that try to explore why SGD generalizes better than Adam. A recent study \cite{simsekli2019tail} observes that the gradient noise in the training phase converges to a heavy-tailed $\alpha$-stable distribution. Another study \cite{zhou2020towards} builds the connection of generalization performance and gradient noise distribution in the training phase. \cite{zhou2020towards} points that the exponential gradient average in Adam, that is $\phi(\cdot)=(1-\beta_1)\sum_{i=1}^{t}\beta_1 (t-i)g_i$, smooths the gradients and diminishes the anisotropic of gradient noise, thus leads to a lighter noise tail than SGD, which damages the generalization performance. 

Inspired by above observations, we propose AdaL. We suggest that conducting the transformation on the current gradient $g_t$ will contribute to the convergences and generalizations. Specifically, we compute the $l_1$-norm of $g_t$. The insight behind this modification is that we need accelerate the convergence speed by amplifying the gradients in the early stage and dampen the oscillation and stabilize the optimization by shrinking the gradient later. In such way, we ensure the consensus of desired optimizers in deep learning. 

Besides, we theoretically analyze the convergence behaviors of AdaL in both convex online optimization and non-convex stochastic online optimization. We rigorously show that the regret  under the convex assumption is upper bounded by $O(\sqrt{T})$, while the convergence rate in the non-convex case is upper bounded by $O(\text{log} T/\sqrt{T})$. It is noticeable that previous proofs are all based on a decreasing $\beta_{1}$ to ensure the convergence, while in our proof we remove this constraint.  We derive a data-dependent regret bound with a constant $\beta_1$ following \cite{alacaoglu2020new}, which bridges the gap between theoretical analysis and practice. Finally, we validate the performance of AdaL on some typical computer vision tasks. In summary, our contribution can be concluded as:
\begin{itemize}
	\item We proposed AdaL, which performs a gradient transformation on Adam without extra parameters. AdaL has two merits: (1) fast convergence as adaptive methods, (2) better generalization performance than Adam.
	\item We have strictly proved the convergence behaviors of AdaL in both convex case and non-convex case with a non-attenuated $\beta_{1}$, which aligns the theoretical proof with the practice. 
	\item We evaluate the empirical performance of AdaL on 
	several benchmarks. AdaL achieves rapid convergence as adaptive methods and good generalization as SGD method. 
\end{itemize} 
\section{Methods}
\subsection{Notations}
We start from the necessary notations in this paper. Let $\theta_i$ denote the $i$-th coordinate of a vector $\theta \in \mathbb{R}^d$. Let $\left \| \theta \right \|_{1}$, $\left \| \theta \right \|_{2}$ and $\left \| \theta \right \|_{\infty}$ denote the $l_1$-norm, $l_2$ norm and $l_{\infty}$-norm, respectively. Given a vector $a\in \mathbb{R}^d$ and a positive definite matrix $M\in \mathbb{R}^{d\times d}$, We denote $M^{-1}a$ as $a/M$ and $\sqrt{M}$ as $M^{1/2}$. Here for a vector,
$\left \| \theta \right \|_{1}=\sum_{i=1}^{d} |\theta_i|$, $\left \| \theta \right \|_{2}=\sqrt{\sum_{i=1}^{d} \theta_i^2}$, $\left \| \theta \right \|_{\infty}=\max_{i} |\theta_i|$. For a matrix $N\in \mathbb{R}^{m\times n}$, we may have a little abuse of notations. We denote $\left \| N \right \|_{1}=\sum_{i=1}^{m} \sum_{j=1}^{n} |N_{ij}|$, which is different from the definition of 1-norm of matrix. The projection operation defined as $P_{\mathcal{X}}^{M}(x)=\arg \min_{y\in \mathcal{X}} \left \|y-x \right \|^2_M$, where $\left \|a \right \|^2_M=\left< a, \text{diag}(M)a \right>$.  We assume $P_{\mathcal{X}}^{M}(x)$ is non-expansive, that is $\left\|P_{\mathcal{X}}^{M}(y)-P_{\mathcal{X}}^{M}(x)\right\|_{M} \leq\|y-x\|_{M}$.
$\mathcal{X}$ has bounded diameter $D$ if $\max_{x,y\in \mathcal{X}} \left \|x-y \right \|_{\infty}\leq D$.

\subsection{AdaL}
Like previous methods, we aim to find an optimization strategy that converges as fast as Adam while generalizes as good as SGD. In this section we develop a new variant of Adam, termed AdaL (Algorithm 1), and provide the convergence analysis.

\begin{algorithm}[h]
	\caption{AdaL} 
	\hspace*{0.02in} {\bf Input:} 
	$x_{1}\in\mathcal{X}$, $\alpha_{t}=\frac{\alpha}{\sqrt{t}}$,
	$\alpha>0,\beta_{1}<1,\beta_{2}<1$
	\\
	\hspace*{0.02in} Set $m_{0}=v_{0}= 0 $ 
	\\
	\hspace*{0.02in} \bf for $t=1$ \bf to $T$ \bf do\\
	\hspace*{0.17in}
	$\begin{array}{l}
		g_{t}=\nabla f_{t}\left(x_{t}\right) \\
		\textcolor{red}{\hat{g_{t}}=\left \| g_t \right \|_{1}g_t}\\  
		m_{t}=\beta_{1} m_{t-1}+\left(1-\beta_{1}\right) \hat{g_{t}} \\
		v_{t}=\beta_{2} v_{t-1}+\left(1-\beta_{2}\right) \hat{g_{t}}^{2} \\
		\textbf{{Bias Correction}} \\
		\qquad \hat{m_t} = \frac{m_t}{1-\beta_{1}^t}, \hat{v_t} = \frac{v_t}{1-\beta_{2}^t}  \\
		\textbf{{Update}} \\
		\qquad x_{t+1}=P_{\mathcal{X}}^{\hat{v}_{t}^{1 / 2}}\left(x_{t}-\alpha_{t} (\hat{v}_{t}^{-1 / 2}+\epsilon) \hat {m_t}\right)
	\end{array}$
\end{algorithm}
The difference from Adam is marked in red. We can see that no extra parameters are introduced in AdaL. The difference between Adam and AdaL is that the latter adds a coefficient to the current gradient $g_t$. Intuitively, we want to accelerate the convergence by amplifying the gradient in the early stage and dampen the oscillation and stabilize the optimization by shrinking the gradient later. 
Note that we choose the $l_1$-norm of $g_t$, that is $\left \| g_t \right \|_{1}$. We make an assumption that $g_t$ in each iterative step is decreasing with oscillation in a general view. We will give specific examples in the following section to illustrate this assumption. 

\subsection{Study on $l_1$-norm of $g_t$}
\begin{figure}[H]
	\label{}
	\centering  
	\subfigure[]{
		\label{a}
		\includegraphics[width=0.25\textwidth]{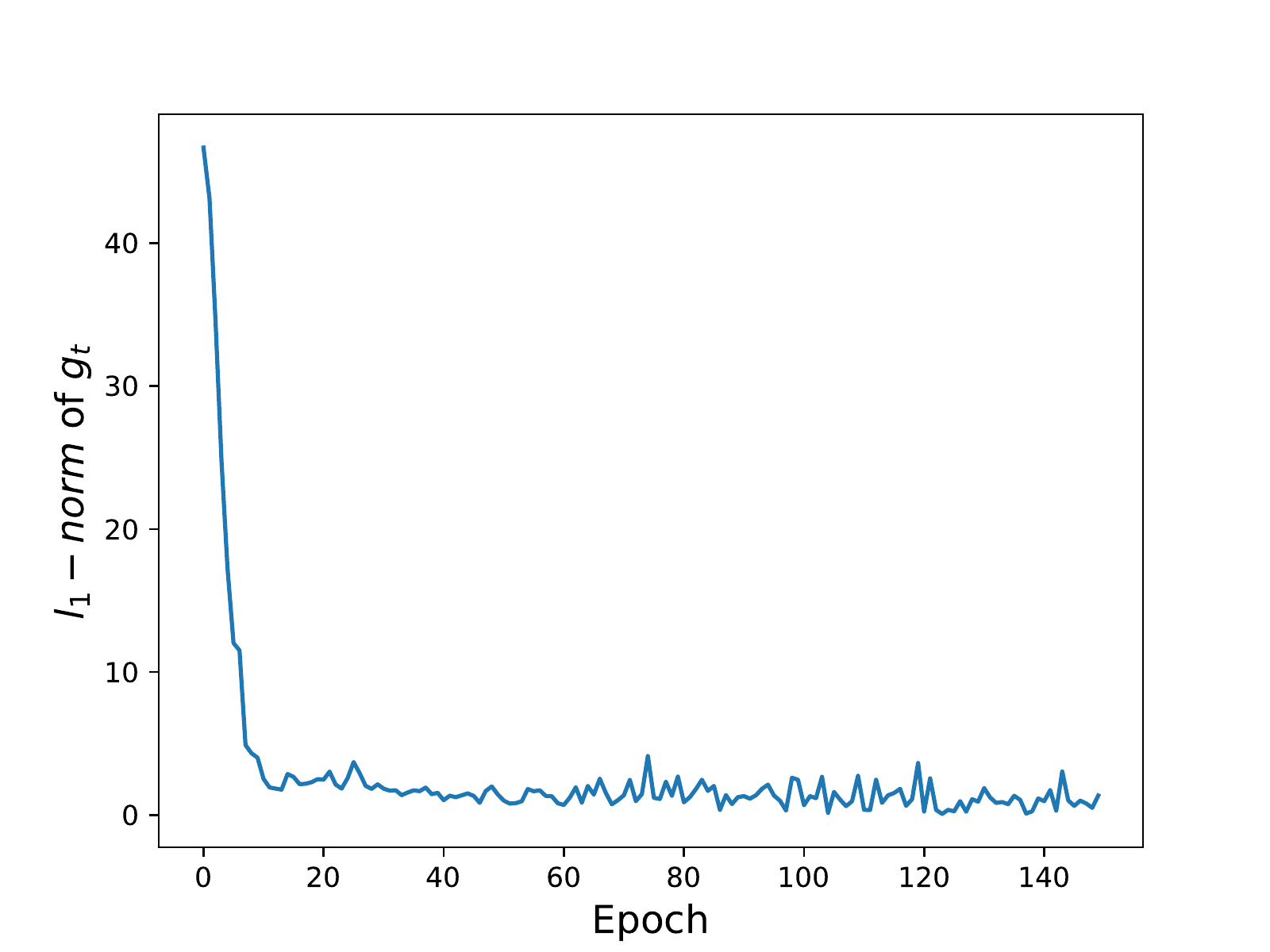}}\hspace{-6mm}
	\subfigure[]{
		\label{b}
		\includegraphics[width=0.25\textwidth]{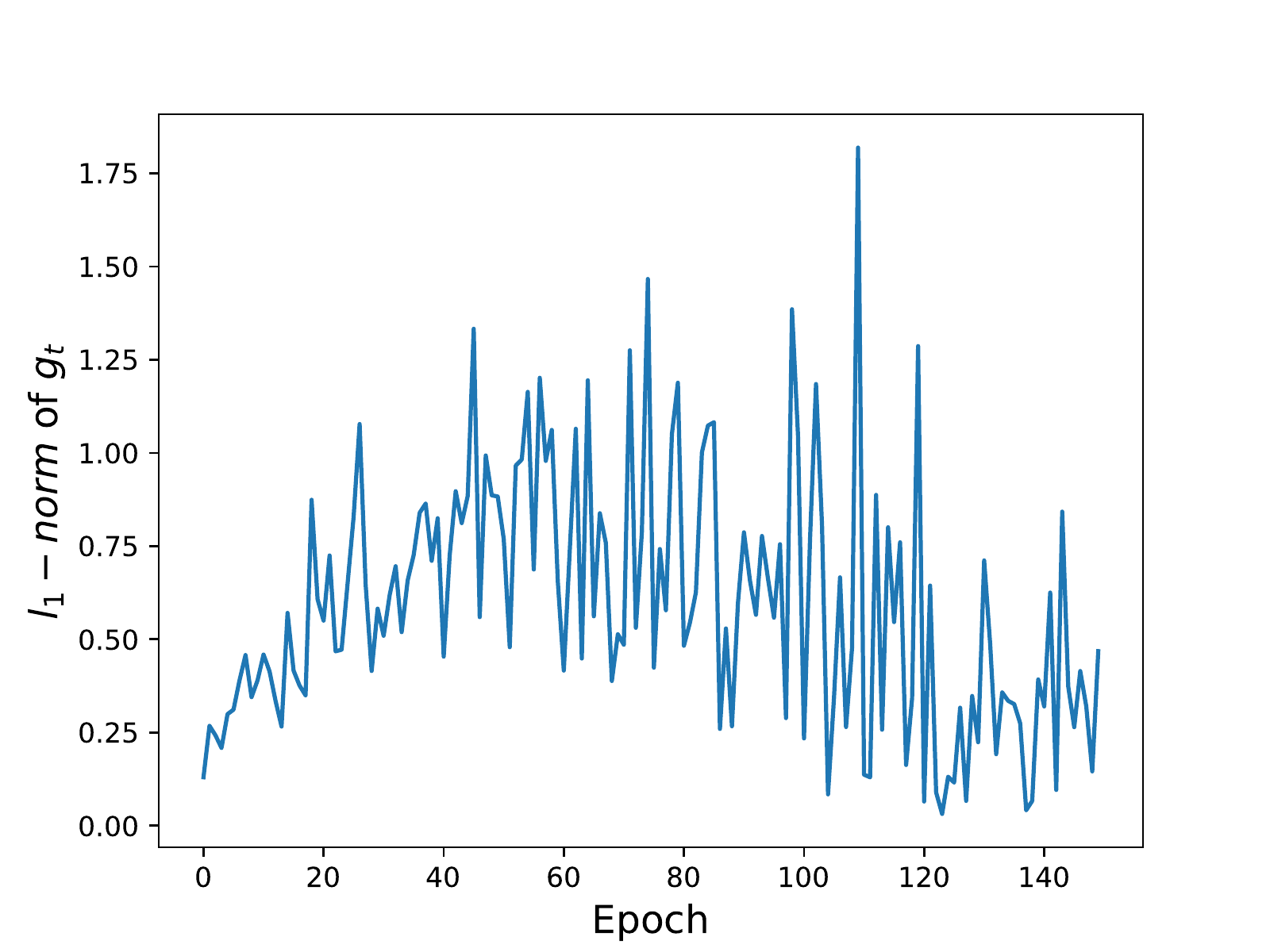}}\\
	\caption{Gradient norm of parameters in ResNet 34 on CIFAR 10 during the training process.}
	\label{cifar10}
\end{figure}

It is noticeable that the insight behind adaptive algorithm is that we choose adaptive step ( determined by the current gradient) to update the parameters in each direction, since frequently occurring and infrequent features should conduct different learning rates. In deep learning models, parameters in the same layer usually share similar properties. The basic unit in the neural network is $AX+b$, where $A$ is weight matrix and $b$ is bias. We compute $\left \| A \right \|_{1}=\sum_{ij} |A_{ij}|$ to amplify or shrink the gradient for each coordinate. All the parameters in the same ``group" share the same coefficient. Here ``group" means the parameters in one layer play a similar role. Actually, it is easy to achieve this modification in PyTorch \cite{paszke2019pytorch} within several lines of code as following:
\lstset{language=Python}
\begin{lstlisting}
	import torch
	
	...
	grad_norm = torch.norm(grad, p=1, dim=None)
	grad_new = grad_norm * grad
	# replace grad with grad_new 
	...
\end{lstlisting}
\begin{figure}[h]{
		\includegraphics[width=0.5\textwidth]{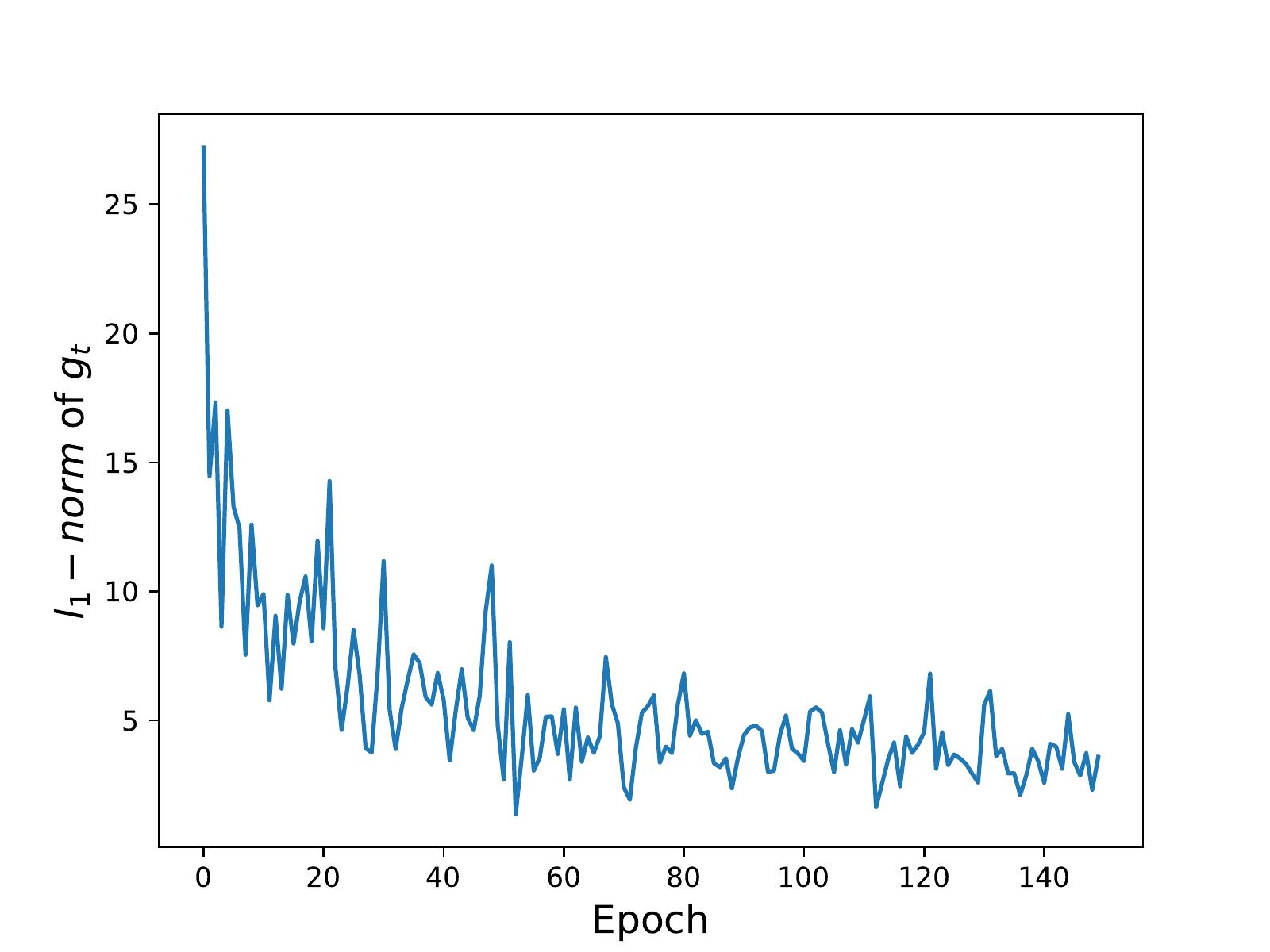}}
	\caption{ Gradient norm of one parameter in MLP on MNIST during the training process.}
	\label{mnist}
\end{figure}
We give two specific examples of $\left \| g_t \right \|_{1}$: MLP on MNIST and ResNet34 on CIFAR 10 in the training process. In Figure \ref{mnist}, we choose one parameter and compute its gradient norm, which is shown decreasing. As for ResNet on CIFAR 10, we choose two typical gradient paradigms. In Figure \ref{a}, it is similar with Figure \ref{mnist}, which is shown decreasing. In Figure \ref{b}, the gradient norm is maintained at a low level.

\subsection{Why AdaL generalizes better than Adam}

It is known that adaptive gradient method usually suffers from poor generalization performance than SGD. Inspired by recent studies \cite{simsekli2019tail} on the stochastic gradient noise in deep neural networks, \cite{zhou2020towards} explains this issue by introducing the  Lévy-driven stochastic differential equations (SDEs) to explore the escaping time of SDEs from a local basin. Their results show that the Radon measure of the basin and the heaviness of gradient noise influence the escaping time, which is related to the generalization performance positively. 
Specifically, the escaping time depends on the Radon measure of the basin on the landscape positively and the heaviness of gradient noise during the training phase negatively. Adam scales each gradient coordinate adaptively, which results a large Radon measure as well as Adam smooths the current gradient by exponential gradient average, which leads to lighter gradient noise tails.  

\begin{figure}[t]{
		\includegraphics[width=0.48\textwidth]{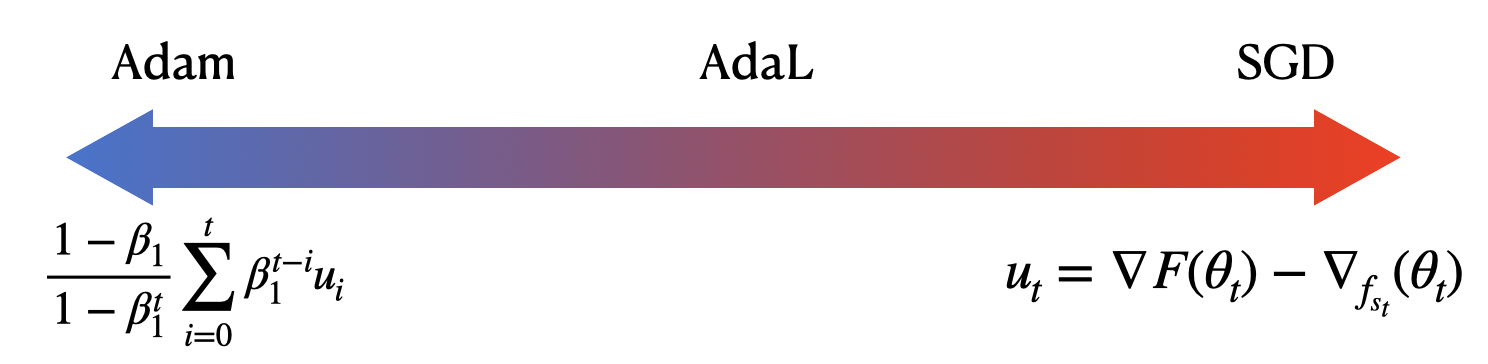}}
	\caption{ Gradient norm of one parameter in MLP on MNIST during the training process.}
	\label{line}
\end{figure}

In Figure \ref{line}, we illustrate the gradient noise of the Adam, SGD and our AdaL in a line. We show that the gradient noise of AdaL is much closer to SGD.

We denote the noise of gradient in optimizers as $u_t$ following \cite{zhou2020towards}:
$$
u_t=\nabla F(\theta_t)-\nabla_{f_{\mathcal{S}_{t}}(\theta_t)} 
$$
where $F$ means the average loss, which is defined as $\min _{{\theta}_t \in {R}^{d}} {F}({\theta_t})=\frac{1}{n} \sum_{i=1}^{n} f_{i}({\theta_t})$, where $f_i(\theta_t)$ is the loss of $i$-th sample.
Accordingly, $\nabla_{f_{\mathcal{S}_{t}}(\theta_t)}$ denotes the gradient on mini-batch $\mathcal{S}_{t}$.\\

Then for popular optimizers SGD and Adam, we have:
\begin{equation*}
	\begin{aligned}
		u^{\text{SGD}}_t &=\sum_{i=0}^{t} g_t - g_t \\
		u^{\text{Adam}}_t &=\frac{1-\beta_1}{1-\beta_1^t}\sum_{i=0}^{t}\beta_1^{t-i}u_t\\
	\end{aligned}
\end{equation*}

For AdaL, 
\begin{gather*}
	u'_t=\nabla F'(\theta_t)-\nabla _{f'_{\mathcal{S}_t}}(\theta_t) 
	=\sum_{i=0}^{t} \left \| g_t \right \|_{1}g_t - \left \| g_t \right \|_{1}g_t \\
	u^{\text{AdaL}}_t=\frac{1-\beta_1}{1-\beta_1^t}\sum_{i=0}^{t}\beta_1^{t-i}u'_t\\
\end{gather*}

In AdaL, the current gradient is amplified by the $l_1$ norm, which alleviates the heaviness of gradient noise. In this way, AdaL reduces the generalization gap with SGD. We refer readers to \cite{zhou2020towards} for more details of gradient noise and Radon measure.

\subsection{Convergence analysis}
\subparagraph{Online Optimization}
At each time step $t$, the optimization algorithm picks a point $x_t$ in a feasible set $\mathcal{X}\in \mathbb{R}^{d}$. Let $f_t$ be the loss function corresponding to the underlying minibatch, and the algorithm incurs loss $f_t(x_t)$. We use regret, which is defined as the sum of all the differences between online prediction $f_t(x_t)$
and loss incurred by the fixed parameter point in $\mathcal{X}$ for all the steps, to evaluate the algorithm.

\begin{equation} \label{regret}
	R(T) = \sum_{t=1}^{T} f_t(x_t) - \min_{x\in \cal X} \sum_{t=1}^{T} f(x)
\end{equation}

\begin{theorem}[Convergence in convex optimization] \label{theorem1}
	\notag
	Let \{$x_t$\} and \{$v_t$\} be the sequences defined in Algorithm 1, $\alpha_{t}=\alpha/{\sqrt{t}}$, $\beta_1<1$, $\beta_2<1$,$\gamma=\frac{\beta_{1}^{2}}{\beta_{2}}<1$. Let $x\in \mathcal{X}$, where $\mathcal{X} \in \mathbb{R}^{d}$ is a compact convex set and has bounded diameter $D$. Let $G=\max \left \| g_t \right \|_{\infty}$  and $M=\max \left \| g_t \right \|_{1}$ for all $t$, then for \{$x_t$\} generated using AdaL, we have the following bound on the regret
	\begin{multline}
R(T) \leq \frac{D^{2} \sqrt{T}}{2 \alpha\left(1-\beta_{1}\right)} \sum_{i=1}^{d} \hat{v}_{T, i}^{1 / 2}+ \\ \frac{\alpha \sqrt{1+\log T}M}{\sqrt{\left(1-\beta_{2}\right)(1-\gamma)}} \sum_{i=1}^{d} \sqrt{\sum_{t=1}^{T} {g}_{t, i}^{2}}
	\end{multline}
\end{theorem}
It is shown that the regret of AdaL is upper bounded by $O(\sqrt{T})$. It is noticeable that we don't need a decaying $\beta_{1t}=\beta_1/t$ to ensure a regret of $O(\sqrt{T})$ like previous proofs. Here we adopt the new regret analysis for Adam-type algorithms provided by \cite{alacaoglu2020new}.

\begin{theorem}[Convergence in non-convex optimization] \label{theorem2}
	\notag
	Under the following assumptions:
	\begin{itemize}
\item $f$ is L-smooth: $\left \|  \nabla f(x)-\nabla f(y) \right \|\leq L\left \| x-y \right \| $, $\forall x,y$.
\item $G=\max_{t} \left \| \nabla f(x_t) \right \|_{\infty}$ and $f_t(x)=f(x, \xi_t)$.
\item $M=\max_t \left \| \nabla f(x_t) \right \|_{1}$.
\item $x_{*}\in \arg \min_x f(x)$ exits.
	\end{itemize} 
	$\alpha_{t}=\alpha/{\sqrt{t}}$, $\beta_1<1$, $\beta_2<1$, $\gamma=\frac{\beta_{1}^{2}}{\beta_{2}}<1$. We have the following bound on the regret
	\begin{multline}
	\frac{1}{T} \sum_{t=1}^{T} \mathbb{E}[\|\nabla f(x_{t})\|^{2}] \leq	\frac{1}{\sqrt{T}M}[\frac{G}{\alpha}(f(x_{1})-f(x_{\star})) +\\ \frac{G^{3}M}{(1-\beta_{1})}\|\hat{v}_{0}^{-1 / 2}\|_{1} + 
\frac{G^{3} d}{4 L \alpha(1-\beta_{1})} + \\
	\frac{G L d \alpha(1-\beta_{1})(1+\log T)}{(1-\beta_{2})(1-\gamma)}]
	\end{multline}

\end{theorem}
Theorem 2 implies the convergence rate of AdaL in the non-convex case is upper bounded by $O(\log T/\sqrt{T})$, which is similar to other optimizers.

Full proof can be found in the Appendix.

\section{Experiments}
In this section, we evaluate the performance of AdaL with other popular optimizers, including Adam, NosAdam, AMSGrad on several benchmarks. We study the task of multi-class classification on CIFAR 10 and CIFAR 100 using deep convolutional neural networks. The setup for each task is detailed in Table \ref{setup}. 
We demonstrate that AdaL can efficiently perform well compared with the state of the art.


\begin{table}[H]
	\centering
	\begin{tabular}{lll}
		\toprule
		Dataset  & Network type  & Architecture \\ 
		\midrule
		CIFAR10    & Deep Convolutional & ResNet-34 \\
		CIFAR10     & Deep Convolutional & DenseNet-121  \\
		CIFAR100     & Deep Convolutional & ResNet-34 \\
		CIFAR100  &Deep Convolutional & DenseNet-121 \\
		\bottomrule
	\end{tabular}
\caption {Details of the models for experiments.}
\label{setup}
\end{table}

Following the previous setting, we fix $\beta_1$ to be 0.9, $\beta_2$ to be 0.999 for all the Adam-like adaptive methods throughout our experiments. We run 150 epochs for all the experiments. 

For image classification task, we use cross-entropy as loss function and apply weight decay $w_d=5\times10^{-4}$ on the parameters to prevent overfitting. We set the mini-batch size to be 128 and search learning rate among \{0.01, 0.001, 0.0001\}. We employ learning rate decay scheme at epoch 50 and epoch 100 by multiplying 0.1.

We also find that many existing methods don't provide the testing loss curve, which indicates the generalization ability of optimizer to some extent. We provide the test loss curve as another metric besides test accuracy. 

All the experiments are conducted on NVIDIA TITAN RTX 24GB GPU using Pytorch \cite{paszke2019pytorch}. 

\subsection{Synthetic Experiments}
We first give two examples of synthetic non-convex objective landscape, named Rastrigin and Rosenbrock benchmark functions. Rastrigin function has one global minimum in (0.0, 0.0) with lots of local minima surrounding it, while Rosenbrock function has one global minimum in (1.0, 1.0) in a flat valley. These two functions are typical because they show some assumptions of the landscape in deep learning. Some previous studies suggest that the landscape has many local minima, which straps the behaviors of optimizers, while recent studies \cite{choromanska2015loss,wu2017towards,he2019asymmetric} show that the flat minima are more common.

\begin{figure}[H]
	\centering  
	\subfigure[Landscape of Rastrigin Function]{
		\label{Rastrigin_landscape}
		\includegraphics[width=0.24\textwidth]{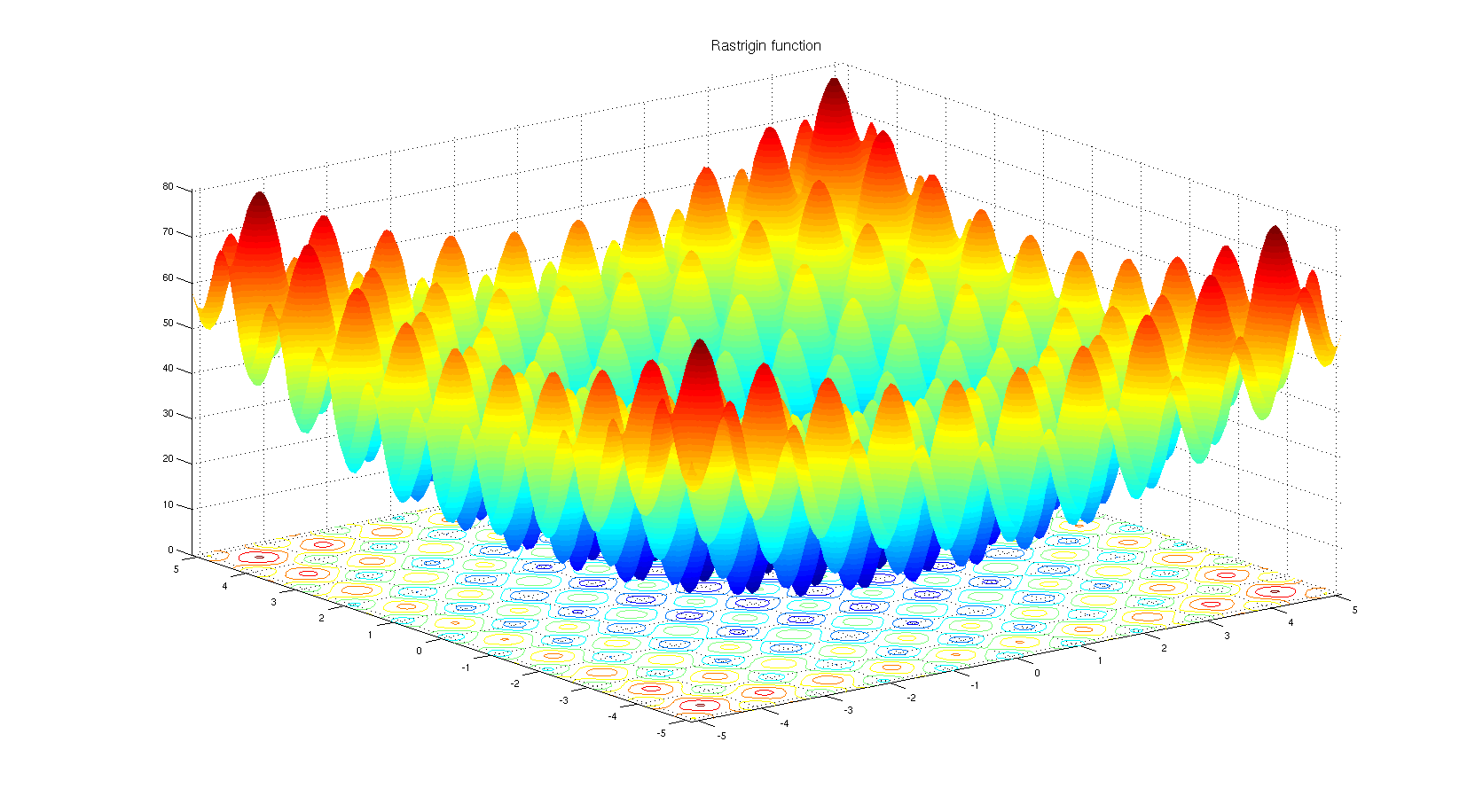}}
	\subfigure[Testing Accuracy]{
		\label{Rastrigin_AdaL}
		\includegraphics[width=0.2\textwidth]{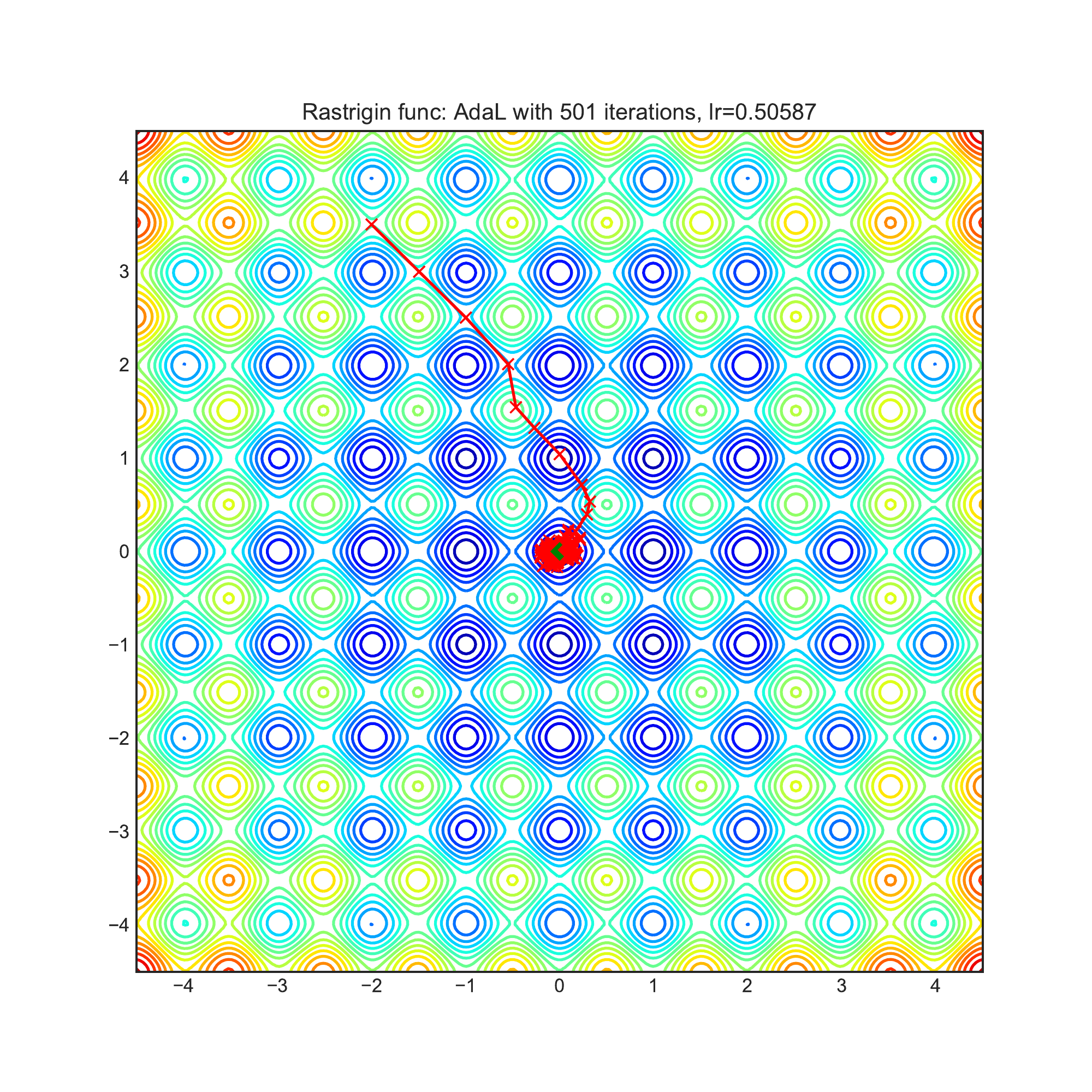}}\\
	\subfigure[Landscape of Rosenbrock Function]{
		\label{Rosenbrock_landscape}
		\includegraphics[width=0.24\textwidth]{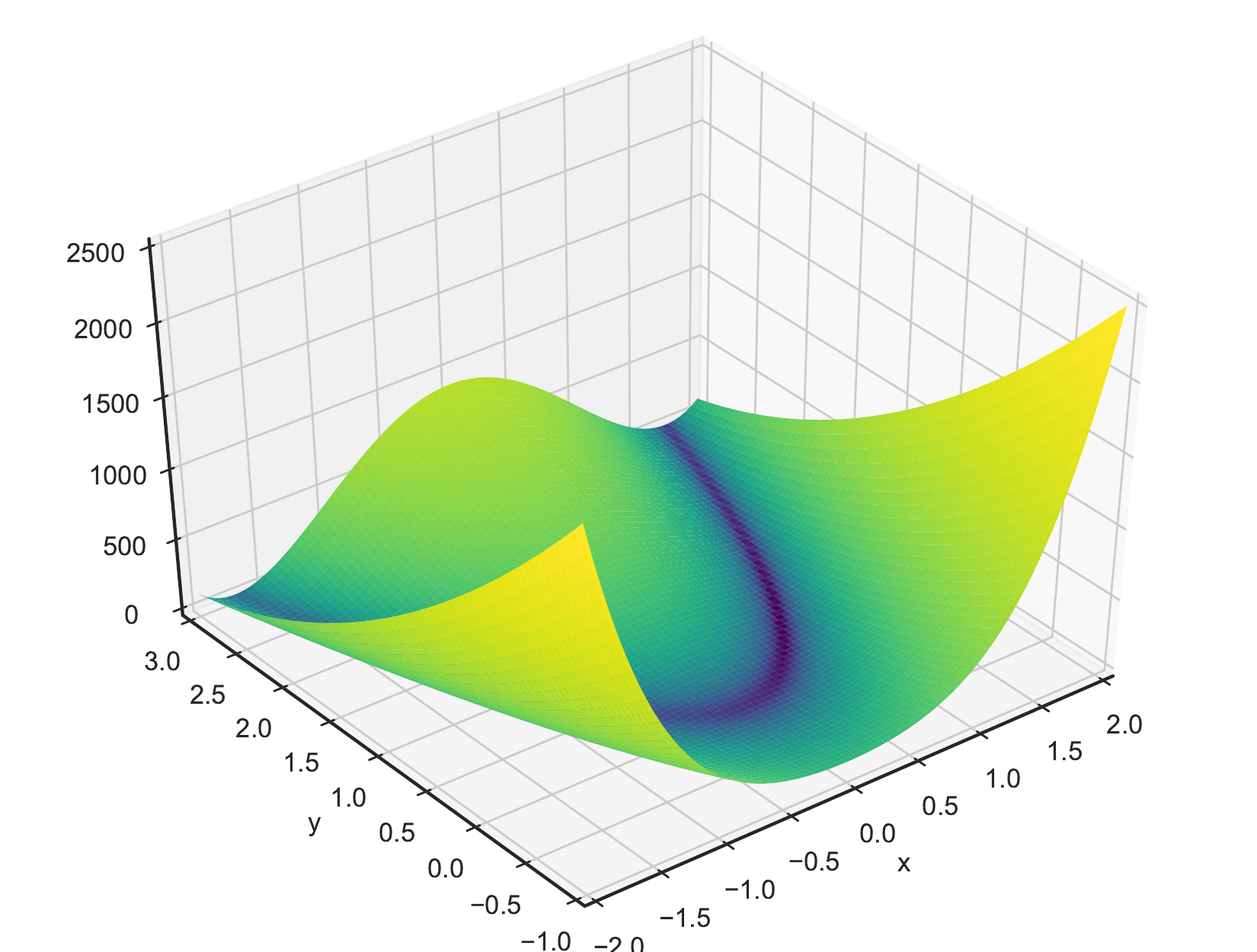}}
	\subfigure[Testing Accuracy]{
		\label{Rosenbrock_AdaL}
		\includegraphics[width=0.2\textwidth]{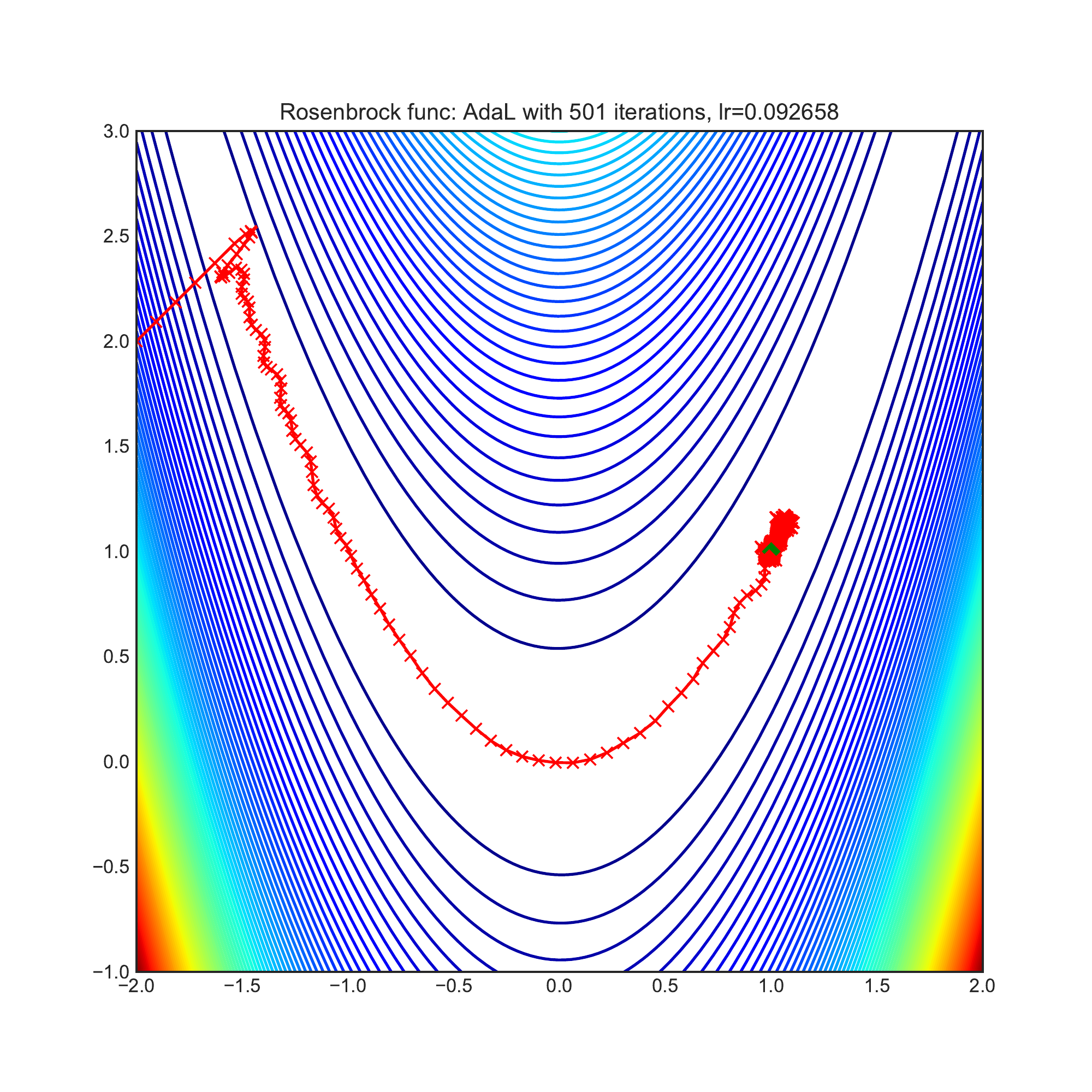}}
	\caption{Rastrigin Function and AdaL's Trajectory}
\end{figure}

\begin{figure*}[htb]
	\centering  
	\subfigure[Log Training Loss]{\hspace{-10mm}
		\includegraphics[width=0.28\textwidth]{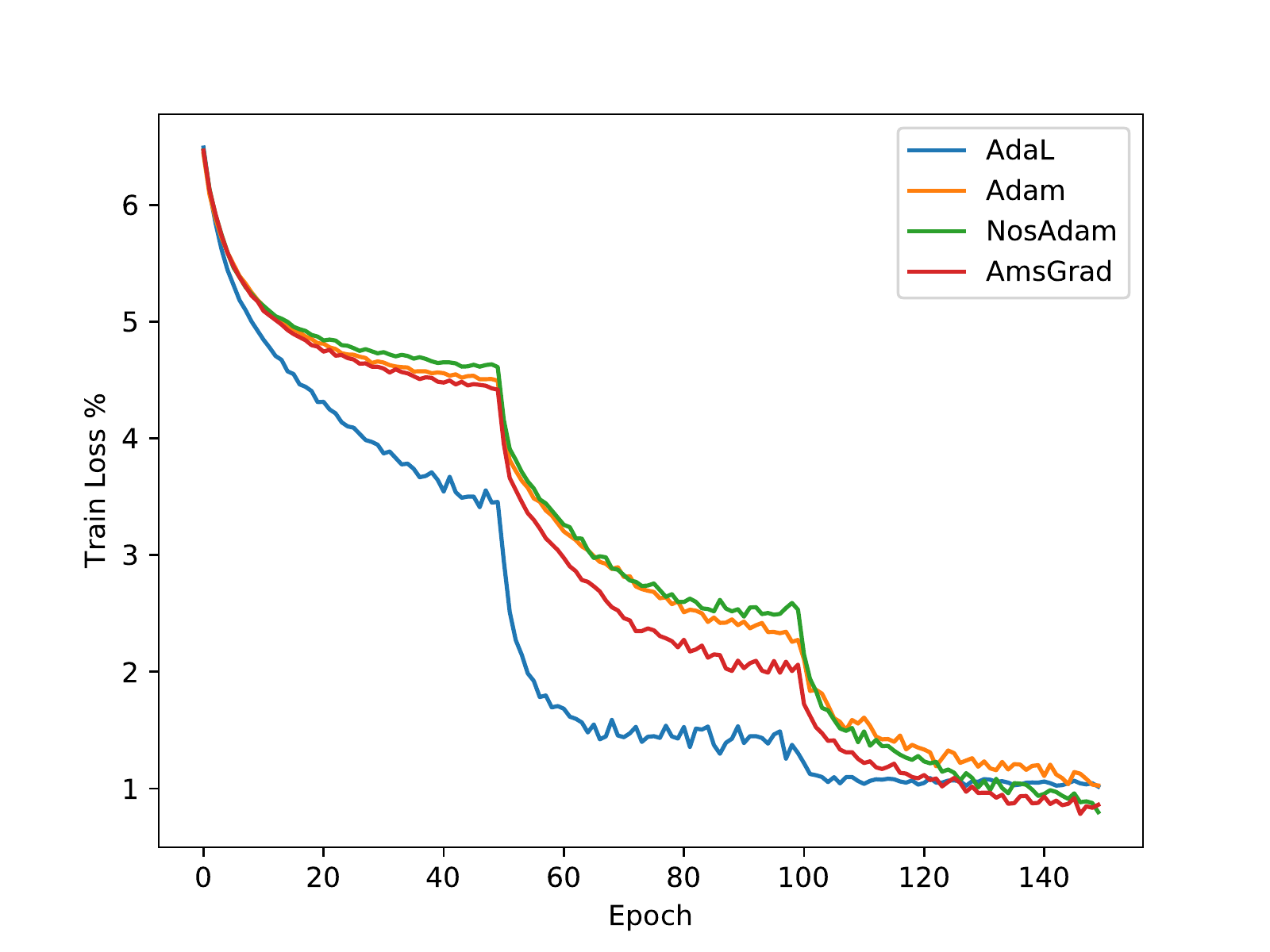}}\hspace{-6mm}
	\subfigure[Training Accuracy]{
		\includegraphics[width=0.28\textwidth]{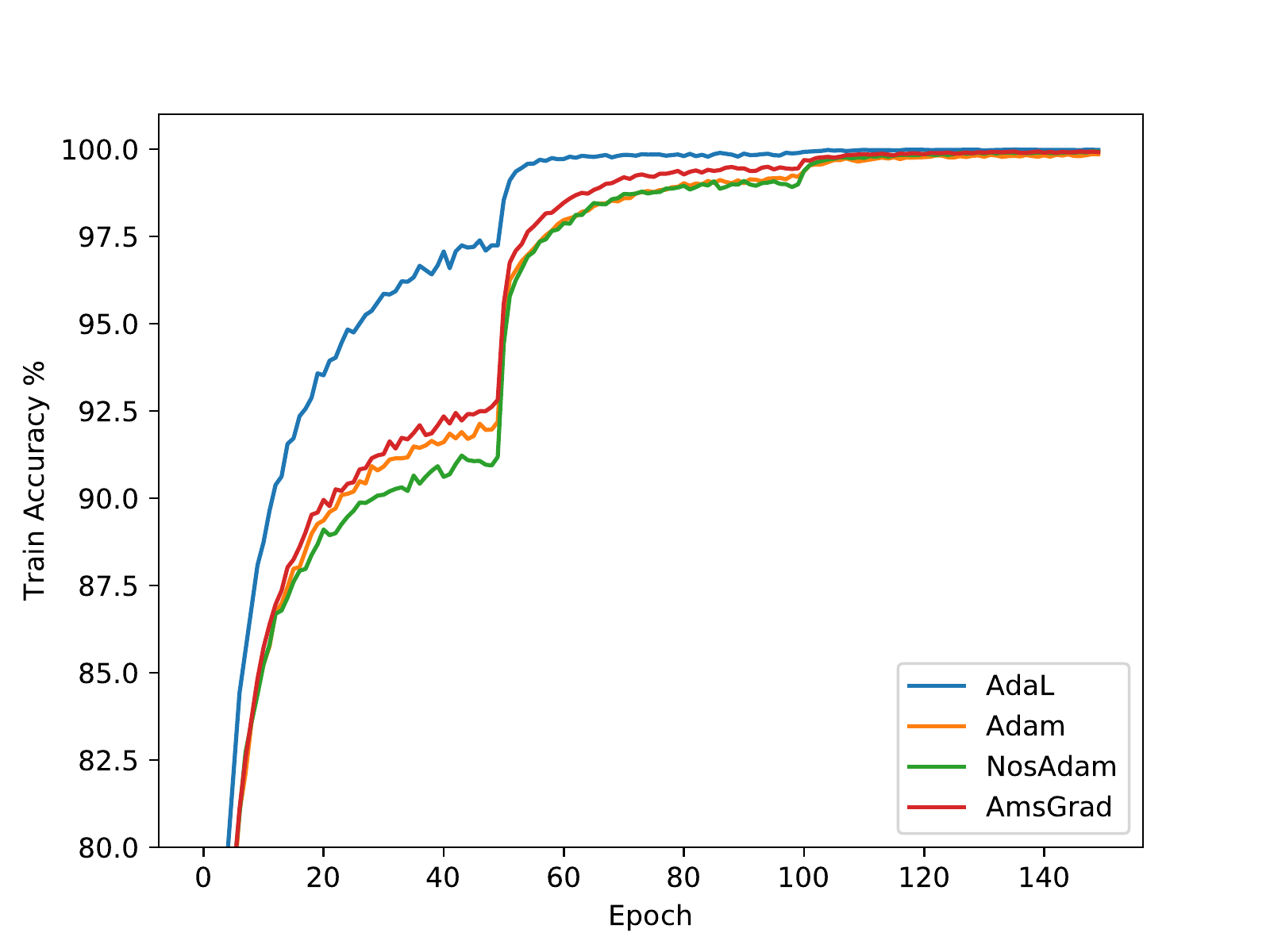}}\hspace{-6mm}
	\subfigure[Log Testing Loss]{
		\includegraphics[width=0.28\textwidth]{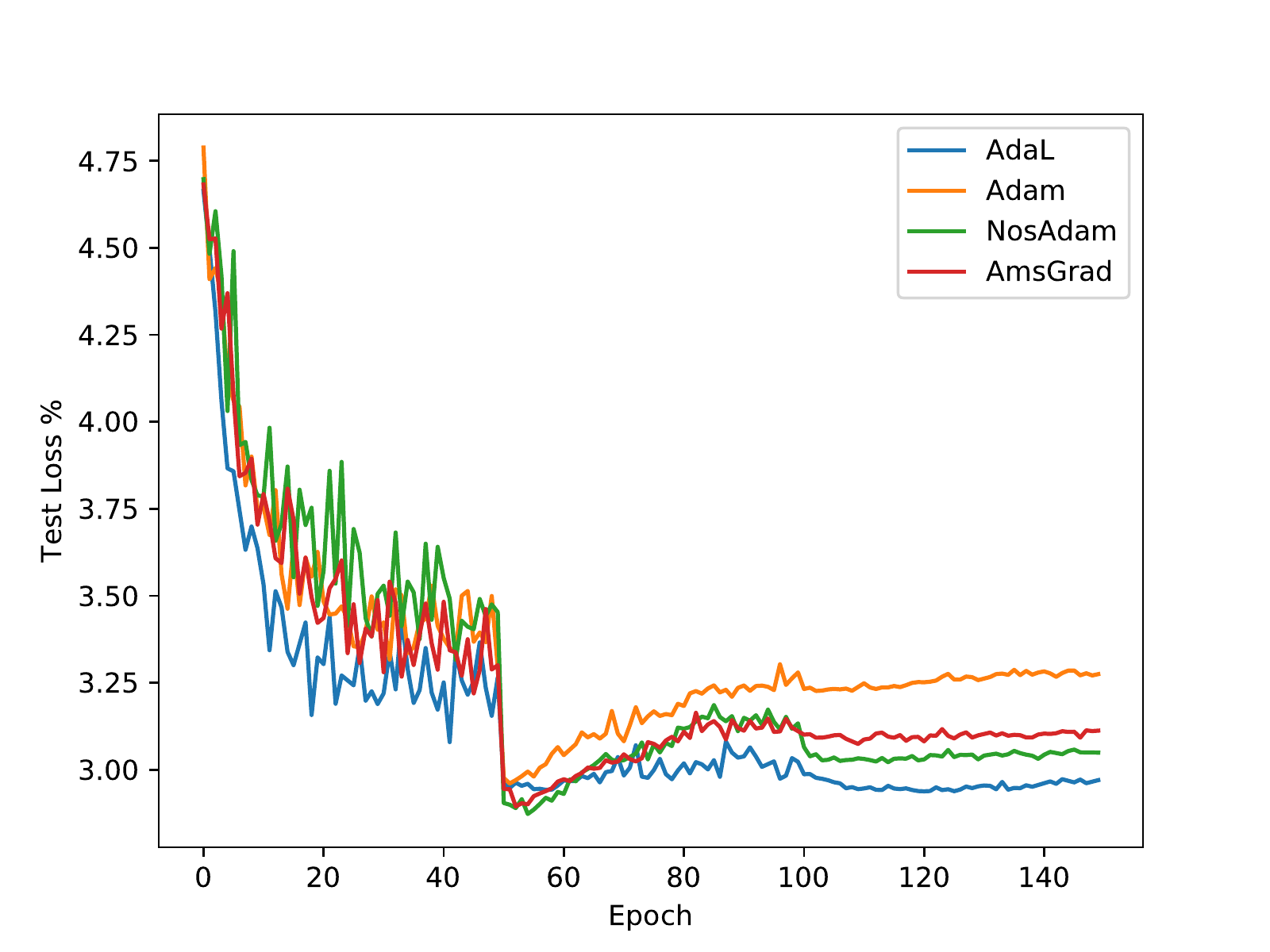}}\hspace{-6mm}
	\subfigure[Testing Accuracy]{
		\includegraphics[width=0.28\textwidth]{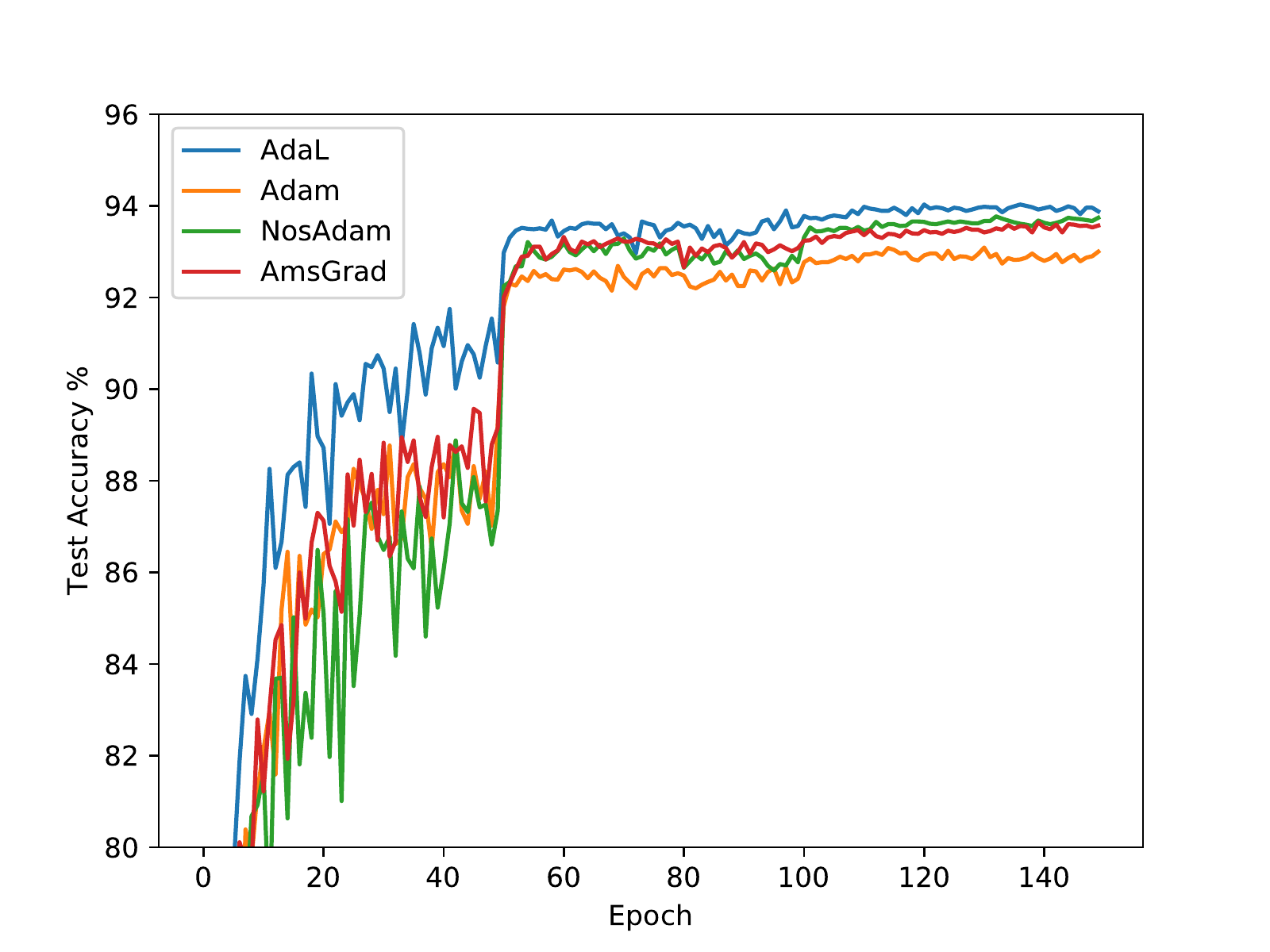}}\hspace{-10mm}
	\caption{Learning curve for ResNet-34 on CIFAR-10.}
	\label{cifar10-resnet}
\end{figure*}

\begin{figure*}[htb]
	\centering  
	\subfigure[Log Training Loss]{\hspace{-10mm}
		\includegraphics[width=0.28\textwidth]{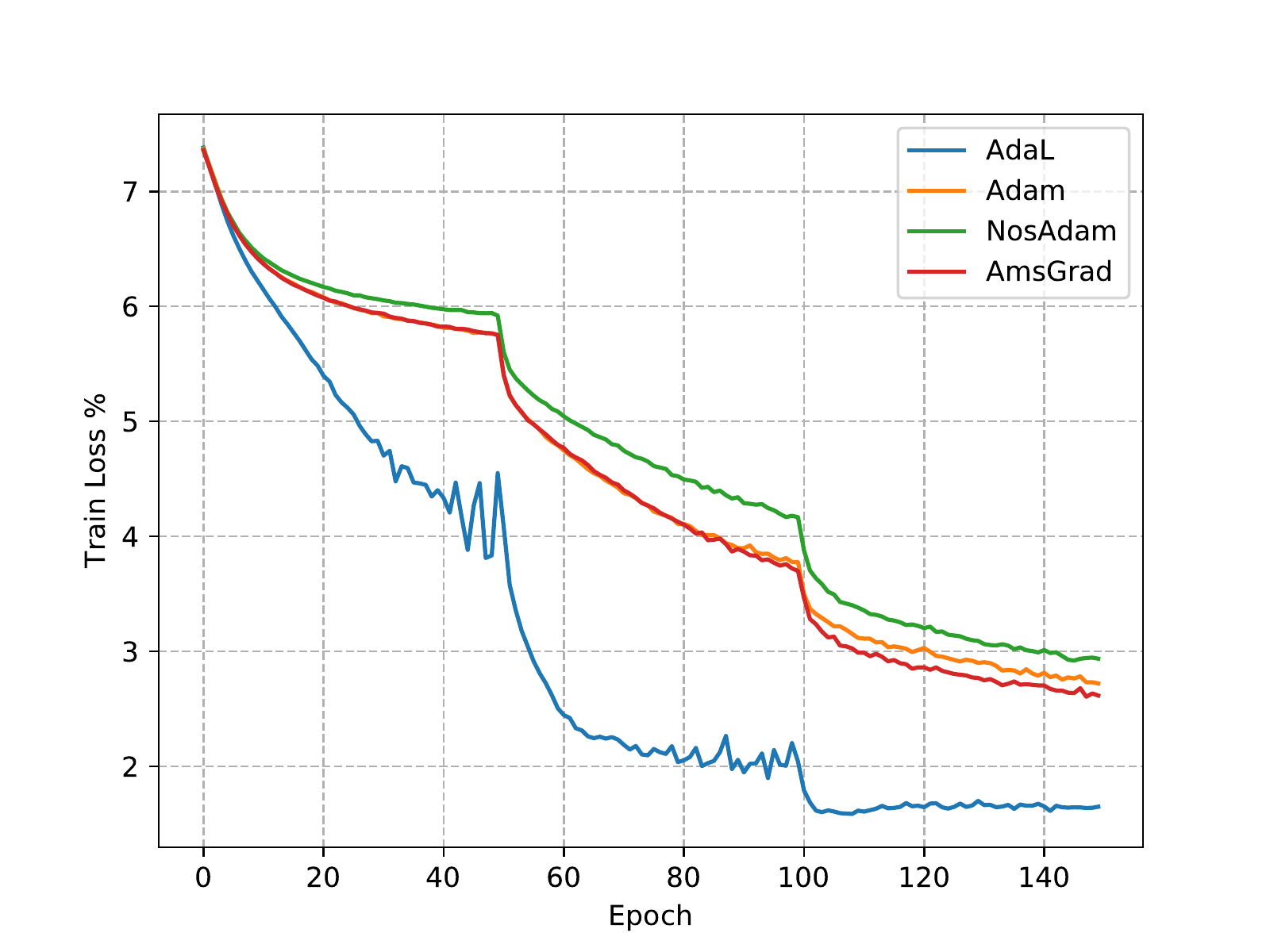}}\hspace{-6mm}
	\subfigure[Training Accuracy]{
		\includegraphics[width=0.28\textwidth]{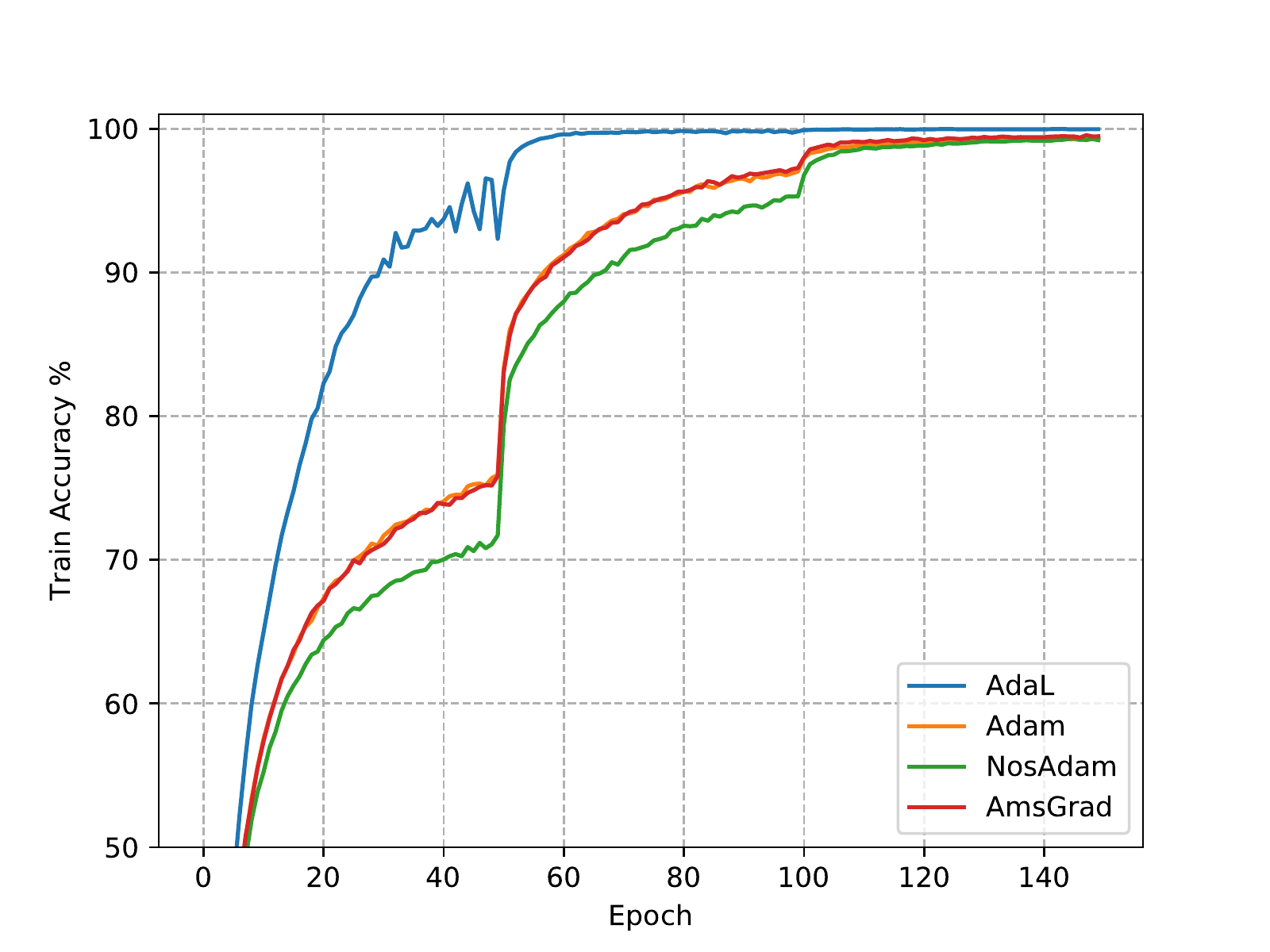}}\hspace{-6mm}
	\subfigure[Log Testing Loss]{
		\includegraphics[width=0.28\textwidth]{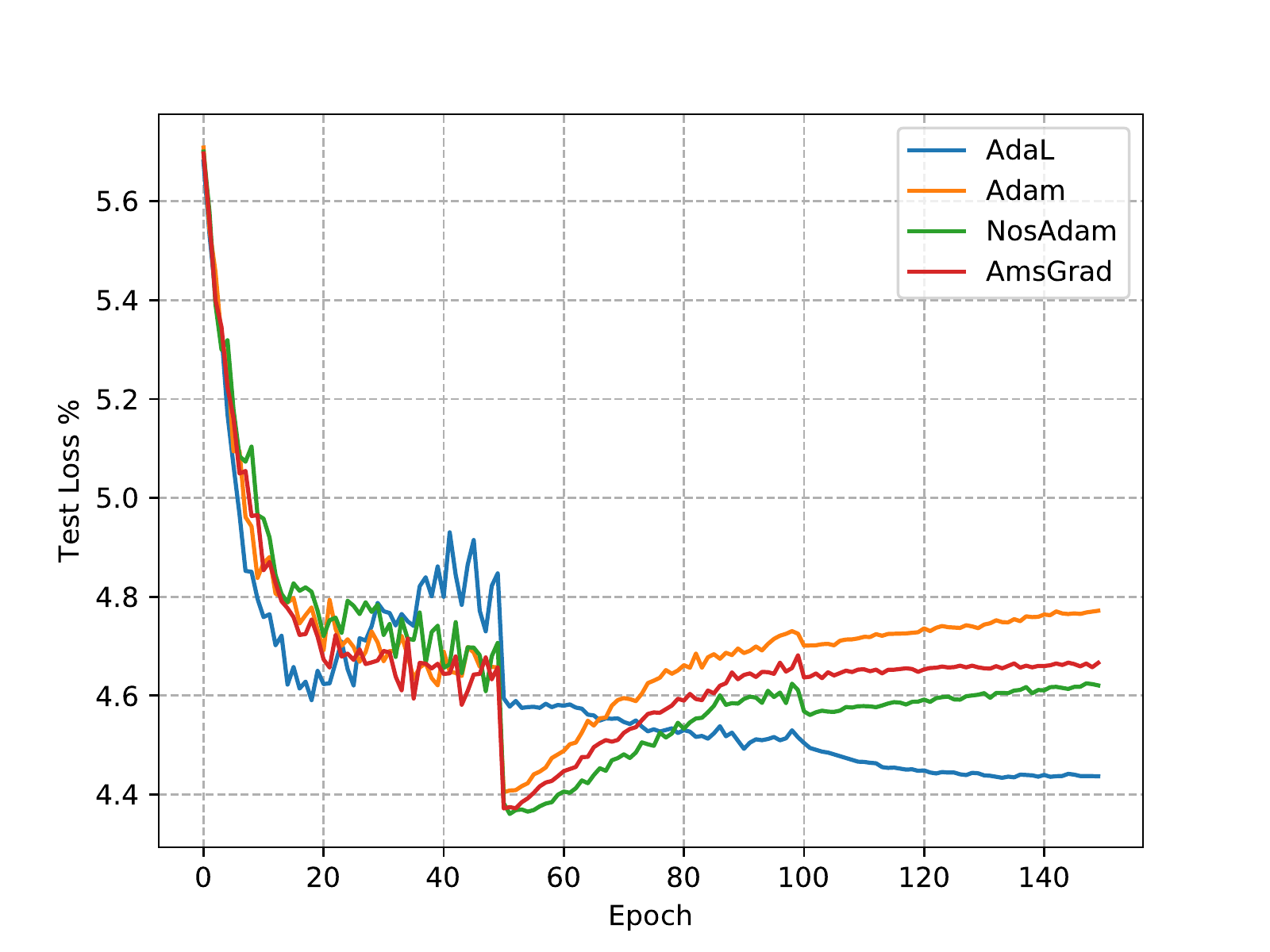}}\hspace{-6mm}
	\subfigure[Testing Accuracy]{
		\includegraphics[width=0.28\textwidth]{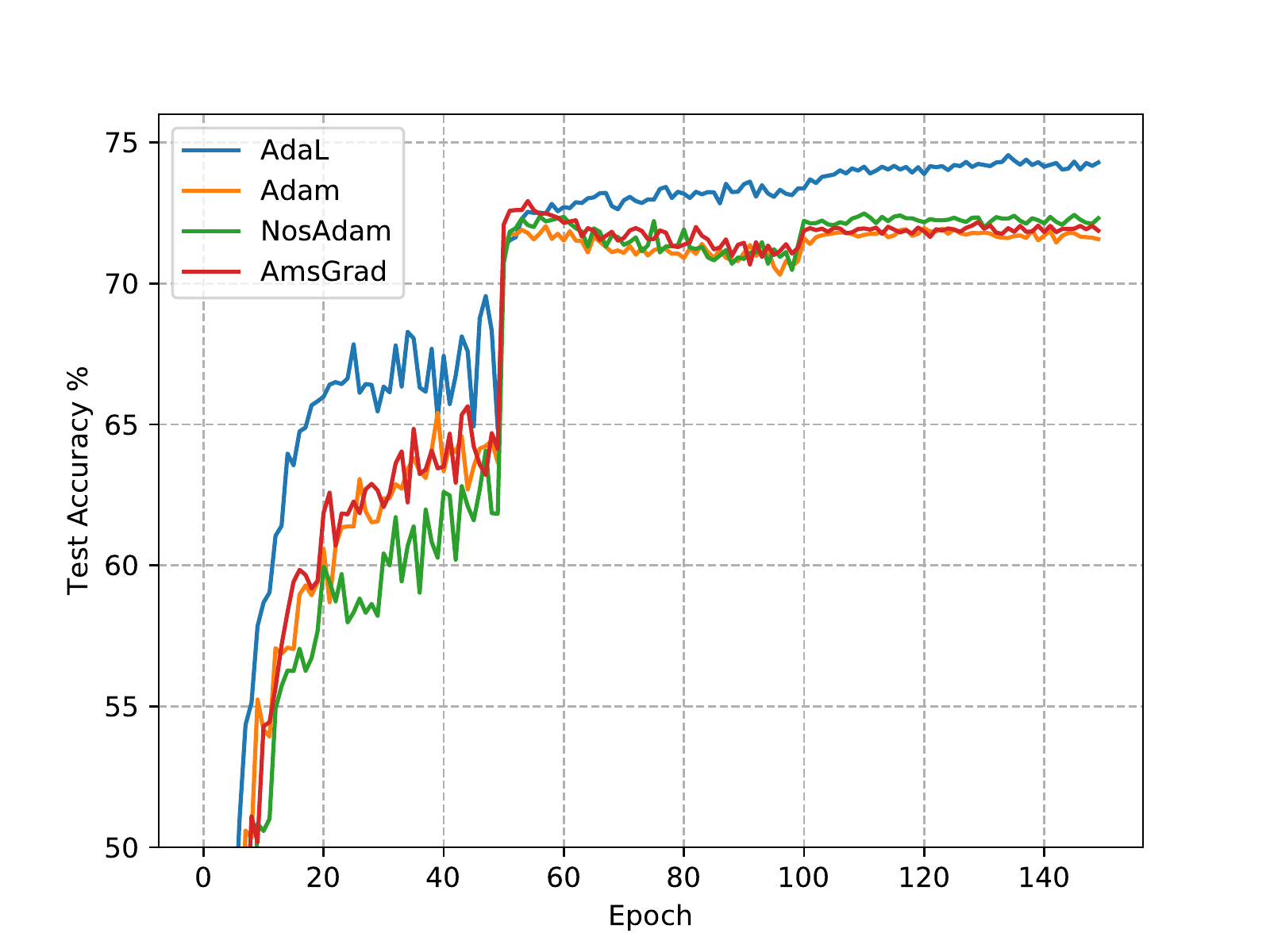}}\hspace{-10mm}
	\caption{Learning curve for ResNet-34 on CIFAR-100.}
	\label{cifar100-resnet}
\end{figure*}


Figure \ref{Rastrigin_landscape} and Figure \ref{Rosenbrock_landscape} show the landscape of the objective functions. Figure \ref{Rastrigin_AdaL} and Figure \ref{Rosenbrock_AdaL} show the  behaviors of AdaL on contour. The green darker indicates the global minimum of Rastrigin function. We conduct the optimization on these two functions and plot the trajectories (red line) of AdaL. Experimental results show that AdaL convergences well on these two situations. As for other optimizers, we refer readers to an open-source GitHub repository \footnote{https://github.com/jettify/pytorch-optimizer} for more details. We can find that AdaL performs well than most optimizers.

%
\subsection{Deep Convolutional Neural Network}
We conduct the task of image classification on CIFAR-10 and CIFAR-100 using two typical networks: ResNet-34 \cite{resnet} and DenseNet-121 \cite{densenet}. For each optimizer, we search for the optimal hyperparameters and report the best parameters.

Different from previous evaluation metric (i.e. accuracy), we consider the loss and accuracy on the testset simultaneously. The intuition is that there exists a gap between test accuracy and test loss, that is, lower loss does not indicate higher accuracy. In the experimental part, we will observe this phenomenon in other baseline optimizers.

The accuracy curves and loss curves are shown in Figure \ref{cifar10-resnet}. We can see that AdaL has fast converge speed in the early stage and achieves higher test accuracy and lower test loss in the final stage. It is noticeable that the test loss of AdaL is also low in several experiments while other optimizers remain a high test loss. Previous studies omit the training and testing loss curves, which are important to reflect the generalization performance to some extent. Experiments illustrate the fast convergence speed and good generalization performance of AdaL.

\begin{figure*}[htb]
	\centering  
	\subfigure[Log Training Loss]{\hspace{-10mm}
		\includegraphics[width=0.28\textwidth]{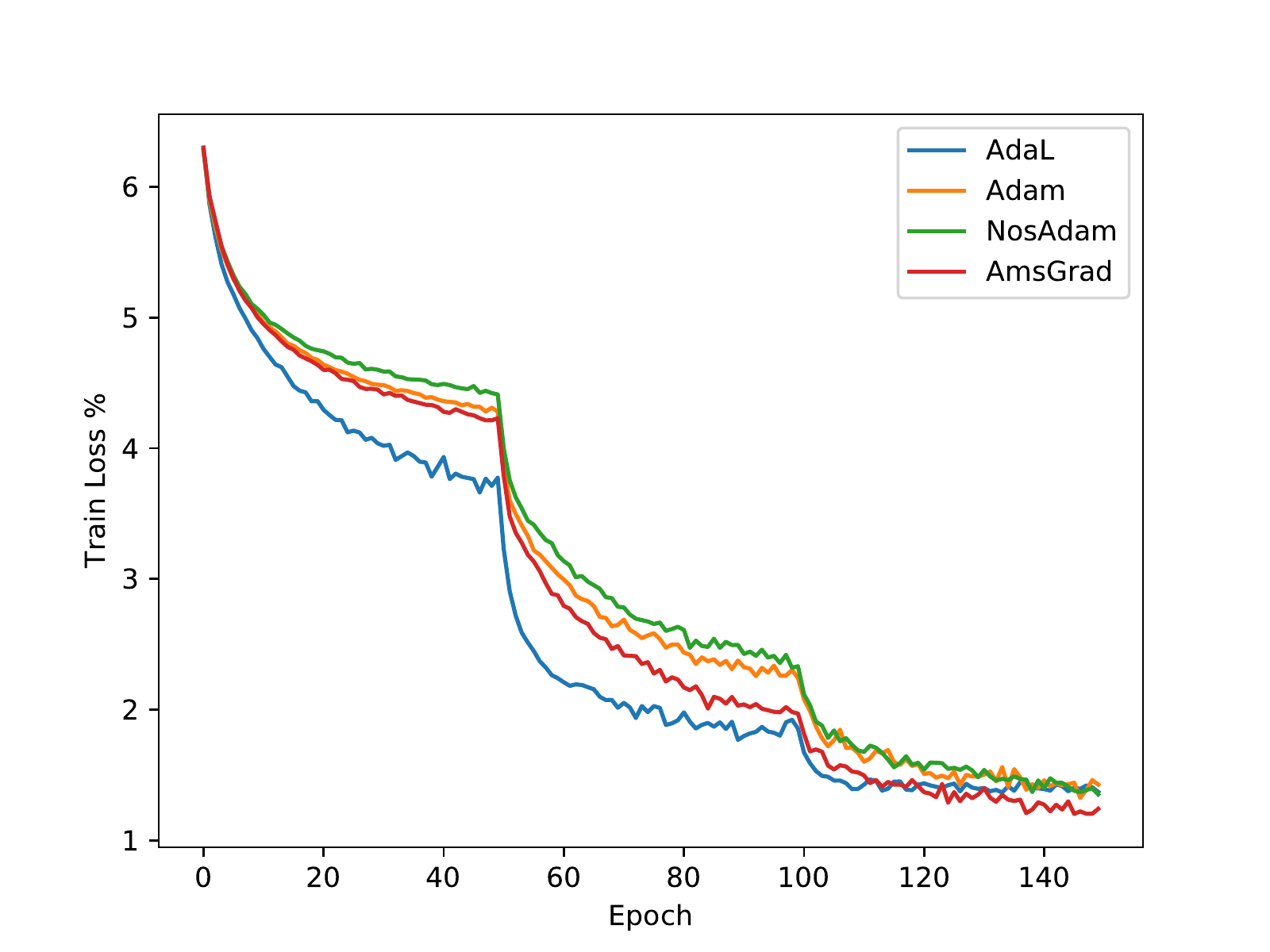}}\hspace{-6mm}
	\subfigure[Training Accuracy]{
		\includegraphics[width=0.28\textwidth]{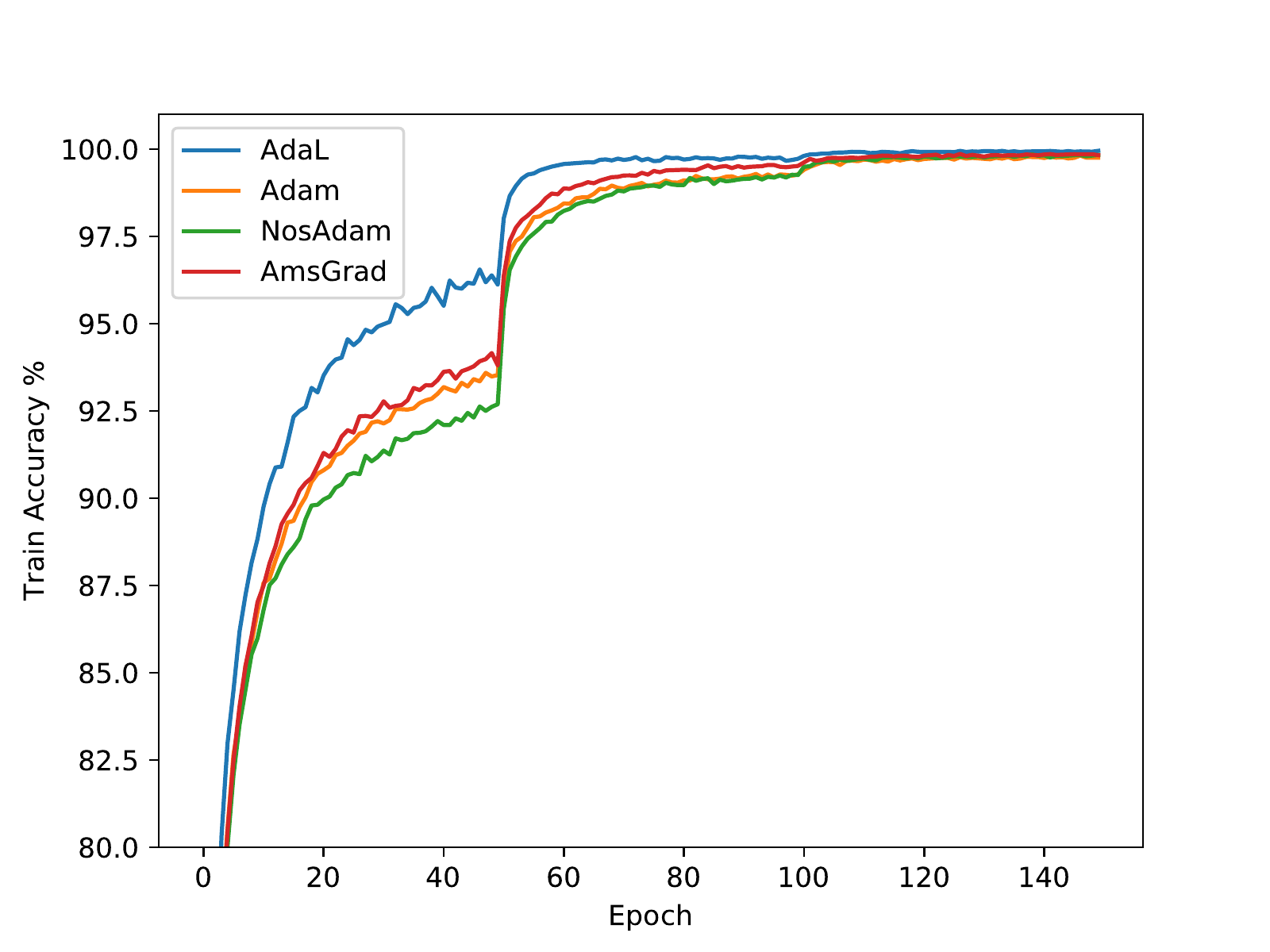}}\hspace{-6mm}
	\subfigure[Log Testing Loss]{
		\includegraphics[width=0.28\textwidth]{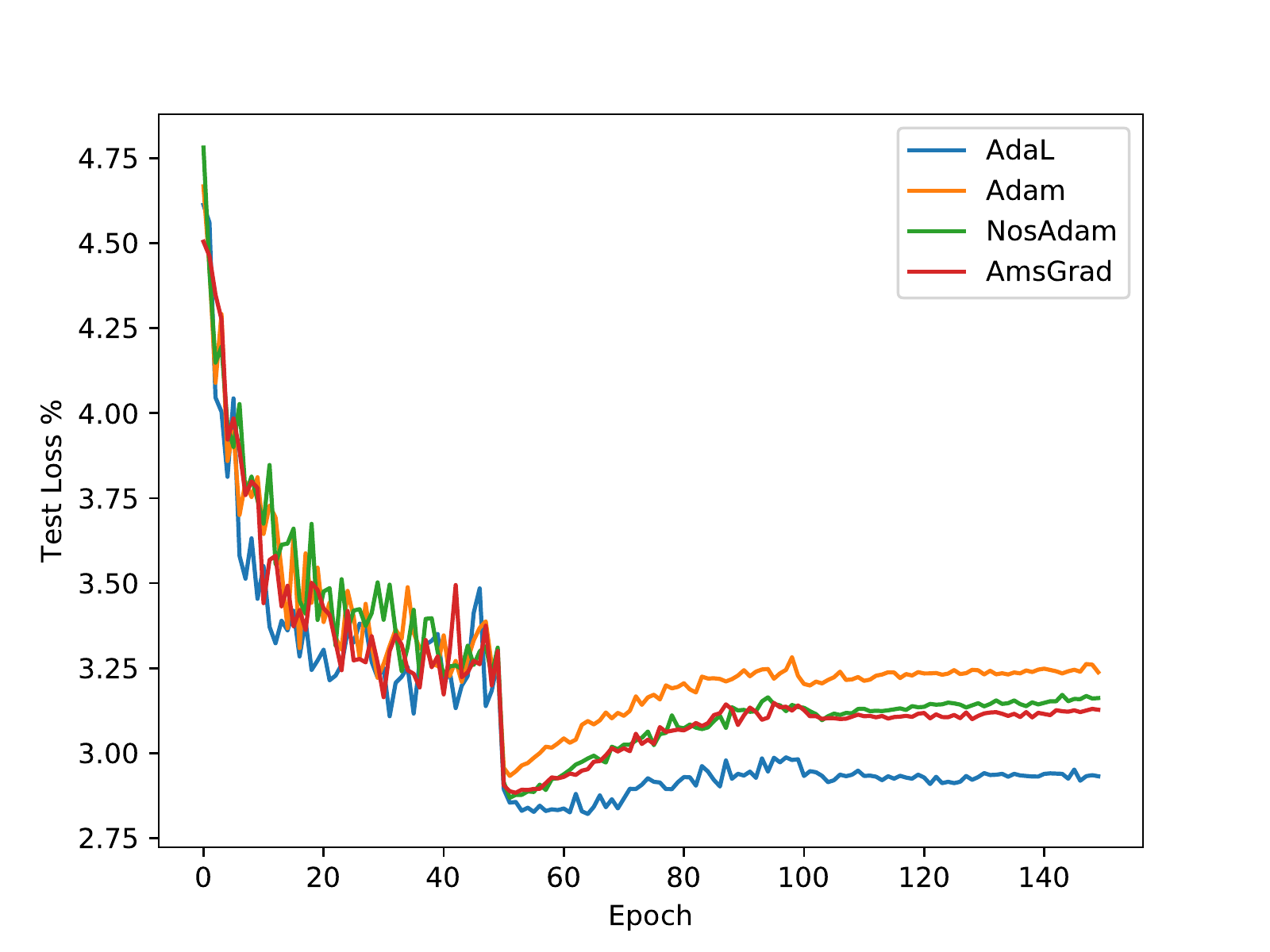}}\hspace{-6mm}
	\subfigure[Testing Accuracy]{
		\includegraphics[width=0.28\textwidth]{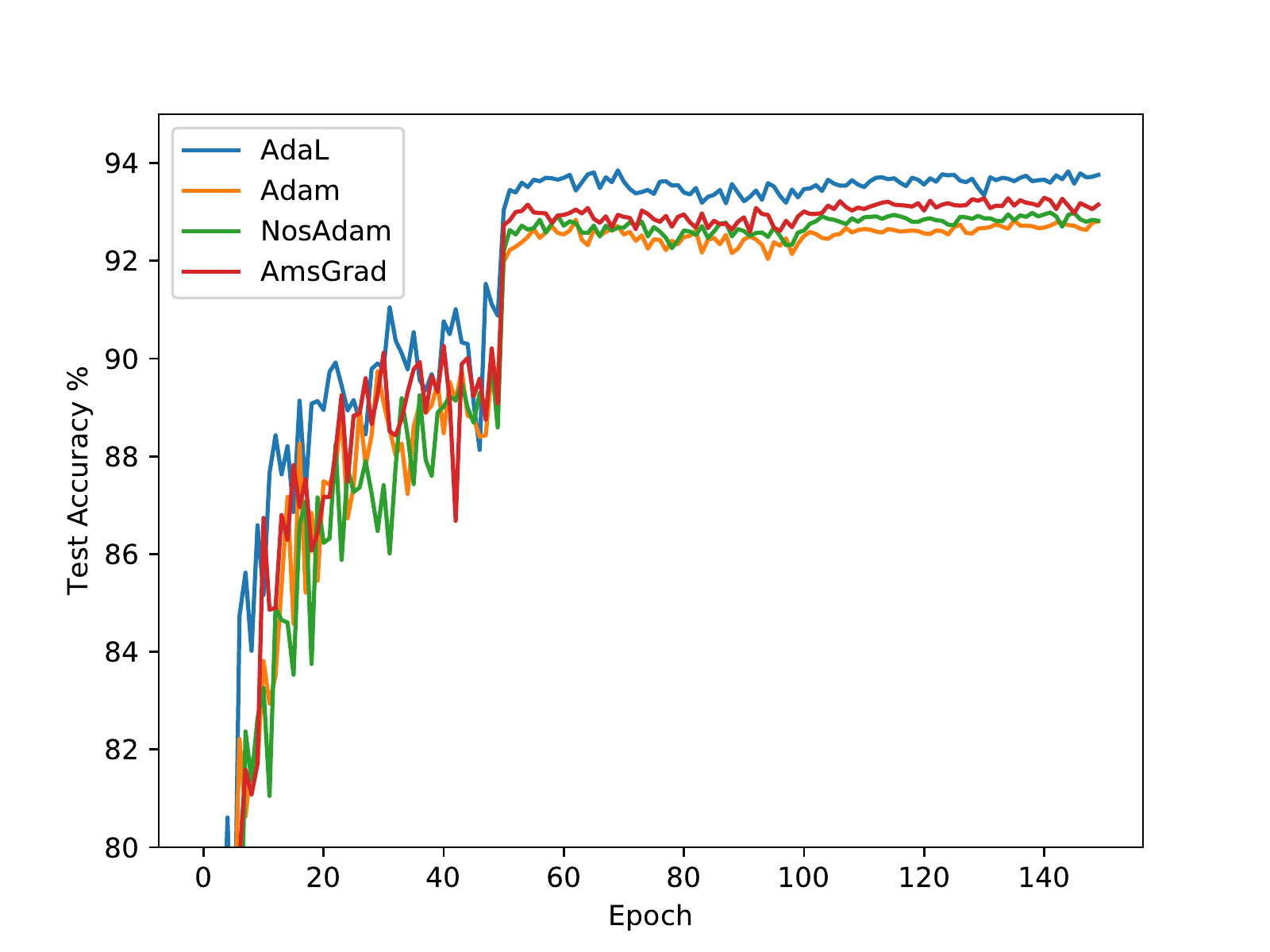}}\hspace{-10mm}
	\caption{Learning curve for DenseNet-121 on CIFAR-10.}
	\label{cifar10-densenet}
\end{figure*}

\begin{figure*}[htbp]
	\centering  
	\subfigure[Log Training Loss]{\hspace{-10mm}
		\includegraphics[width=0.28\textwidth]{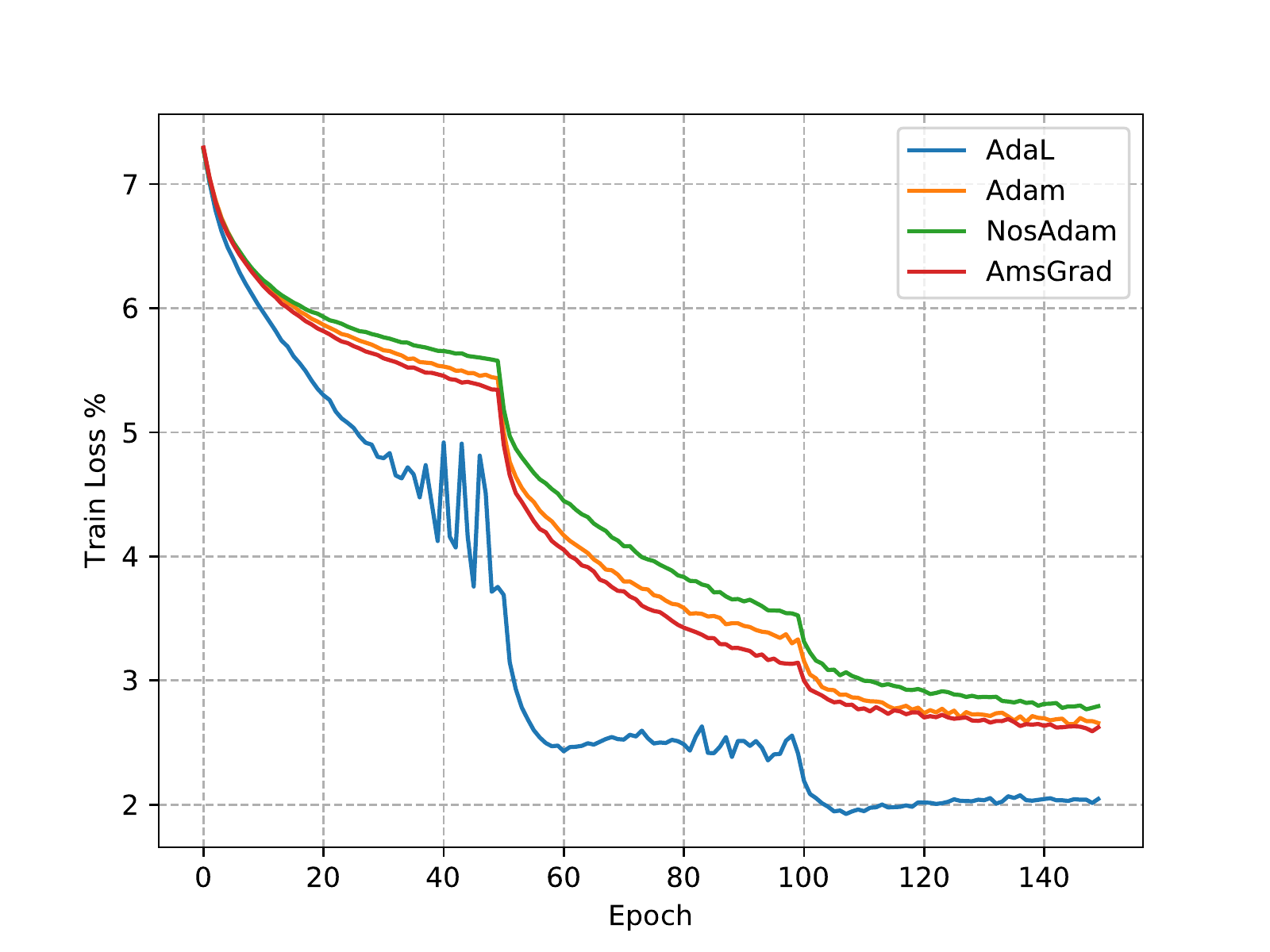}}\hspace{-6mm}
	\subfigure[Training Accuracy]{
		\includegraphics[width=0.28\textwidth]{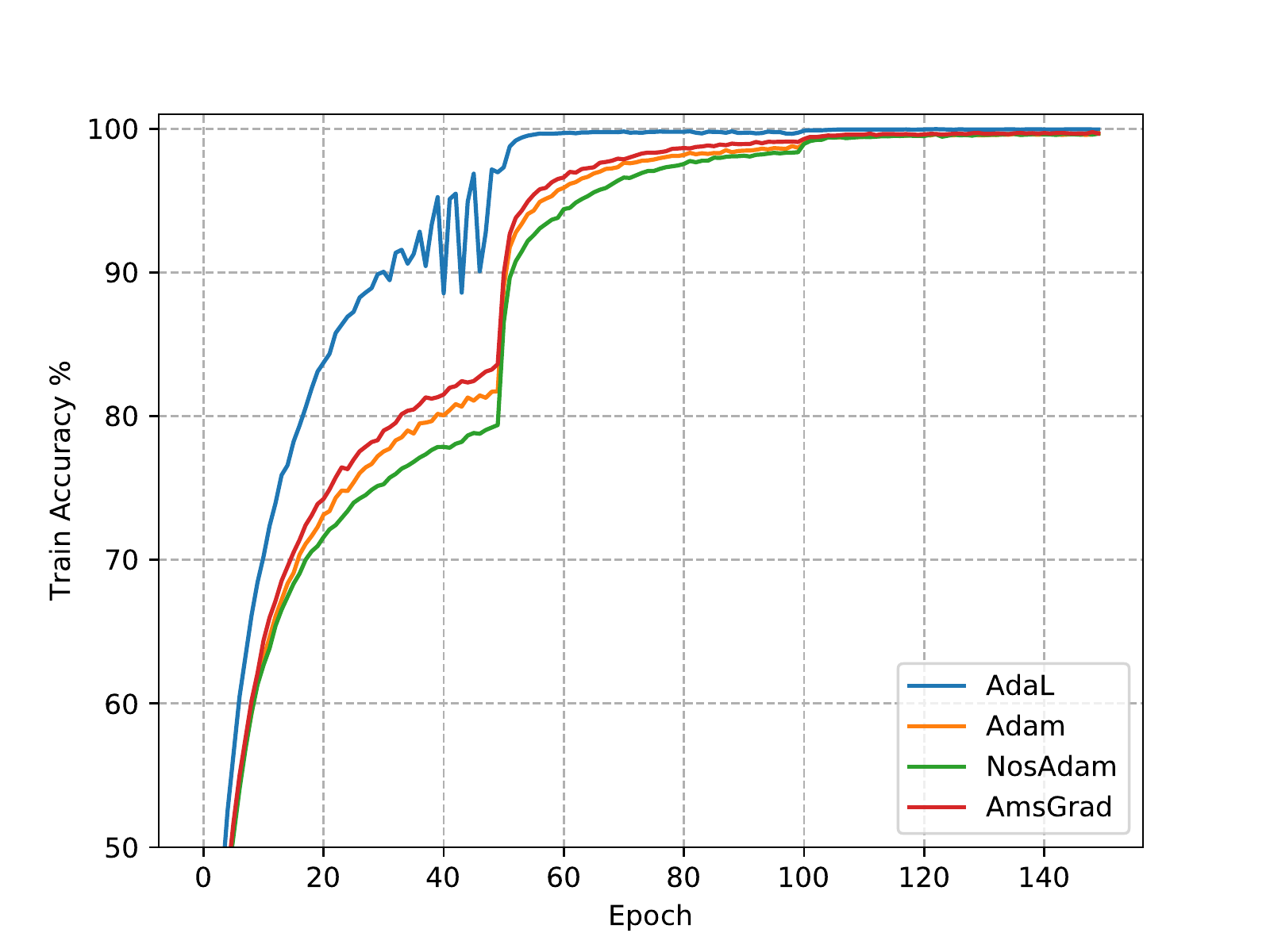}}\hspace{-6mm}
	\subfigure[Log Testing Loss]{
		\includegraphics[width=0.28\textwidth]{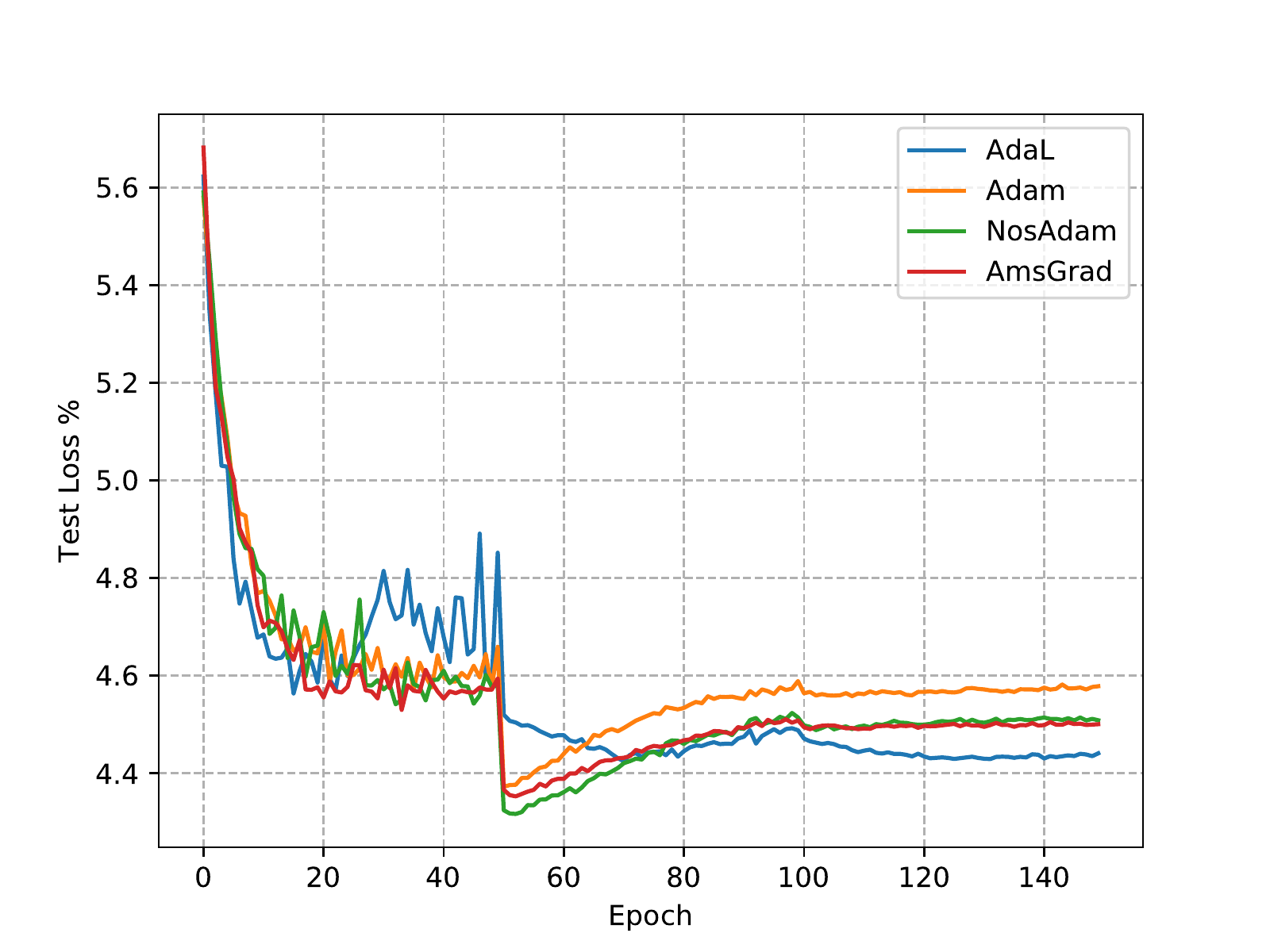}}\hspace{-6mm}
	\subfigure[Testing Accuracy]{
		\includegraphics[width=0.28\textwidth]{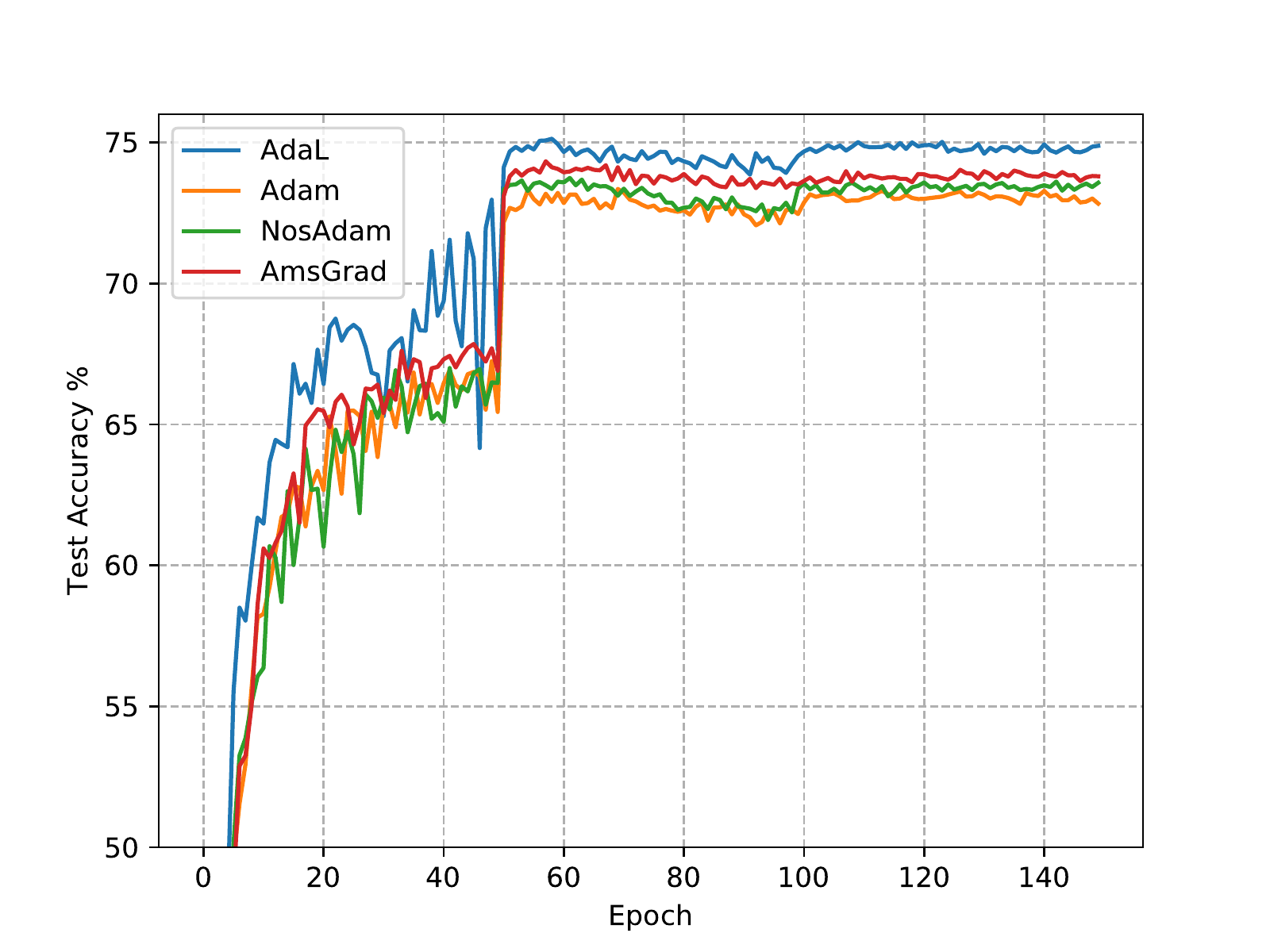}}\hspace{-10mm}
	\caption{Learning curve for DenseNet-121 on CIFAR-100.}
	\label{cifar100-densenet}
\end{figure*}



%

\subsection{Analysis}
To verify the efficacy of AdaL, we conduct several experiments to compare the performance based on the results shown above. It is easy to find that AdaL converges fast compared with baseline methods. It is also noticeable that convolutional and fully connected layers play different parts in deep CNN models. Parameters in different convolutional layers have different roles, which leads to a distinct variation of gradients and their norms.
\section{Future Work}
It is worthwhile to explore several other problems, which is important to understand the optimization and generalization in deep learning. For example, current optimizers always ignore the discussion of generalization from a theoretical view. Many studies aim to explore the global minimum or local minima, which is not the key of deep learning. It is more important to explore the property of optimal point searched by optimizers when designing optimization algorithms.
We should pay more attention to the flat minima or sharp minima, which influences the generalization performance to some extent heavily \cite{hardt2016train,kleinberg2018alternative,xie2020diffusion}. Current optimizers should reduce this fragmentation between optimization and generalization. Besides, the gradient noise plays a vital role in the dynamics of optimization, as the noise is anisotropic and location-dependent, which implies complex dynamics. Last but not least, efficient second-order gradient may provide more insight on the convergence and generalization, which remains to explore.
\section{Conclusion}
In this study, we propose an improved Adam called AdaL, which accelerates the convergence by amplifying the gradient in the early stage, as well as dampens the oscillation and stabilizes the optimization by shrinking the gradient later. Compared with Adam, this modification alleviates the smoothness of gradient noise, which produces better generalization performance. We theoretically proved the convergence of AdaL and empirically demonstrated its powerful global search ability through two complex non-convex functions. Moreover, extensive experiments on image classification problems show that AdaL can effectively accelerate the convergence speed of deep neural networks and improve the generalization ability of their models.


\clearpage
\bibliographystyle{named}
\bibliography{refer}

\end{document}